\documentclass[lettersize,journal]{IEEEtran}
\usepackage{amsmath}
\usepackage{graphicx}
\usepackage{algpseudocode}
\usepackage{algorithm}
\usepackage{booktabs}
\usepackage{cite}
\usepackage{tikz}
\usepackage{float}
\usepackage{multirow}
\usepackage{color}
\usepackage{subfigure}

\usepackage{epstopdf}
\usepackage{hyperref}
\usepackage{longtable}
\usepackage{diagbox}
\usepackage{changepage}
\usepackage{comment}
\usepackage{cases}
\usepackage{makecell}

\usepackage{enumitem}
\usepackage{cuted}
\usepackage{cryptocode} 
\usepackage{fancybox,framed}

\usepackage{lineno,hyperref}
\modulolinenumbers[5]

\usepackage{multirow}
\usepackage{epsfig}
\usepackage{xcolor}

\usepackage{booktabs}

\usepackage{arydshln}
\usepackage{url}

\begin{document}

\title{Deep Learning Approaches for Anti-Money Laundering on Mobile Transactions: Review, Framework, and Directions}
\author{Jiani Fan, Lwin Khin Shar, Ruichen Zhang, Ziyao Liu, Wenzhuo Yang, \newline Dusit Niyato,~\IEEEmembership{Fellow,~IEEE}, Bomin Mao, and Kwok-Yan~Lam,~\IEEEmembership{Senior~Member,~IEEE}
\thanks{Jiani Fan, Ruichen Zhang, Ziyao Liu, Wenzhuo Yang, Dusit Niyato and Kwok-Yan Lam are with the School of Computer Science and Engineering, Nanyang Technological University, Singapore (e-mail: {jiani001, ziyao002}@e.ntu.edu.sg, {ruichen.zhang, wenzhuo.yang, dniyato, kwokyan.lam}@ntu.edu.sg). \textit{(Corresponding author:Jiani Fan)}

Lwin Khin Shar is with the School of Computing and Information Systems, Singapore Management University, Singapore (e-mail: lkshar@smu.edu.sg). Bomin Mao is with the School of Cybersecurity, Northwestern Polytechnical University, China (e-mail: maobomin@nwpu.edu.cn).
}
}

\markboth{IEEE ... JOURNAL, ~VOL.~XX, NO.~XX, FEBRUARY, ~2022}%
{}

\maketitle

\begin{abstract}

Money laundering is a financial crime that obscures the origin of illicit funds, necessitating the development and enforcement of anti-money laundering (AML) policies by governments and organizations. \textcolor{black}{The proliferation of mobile payment platforms and smart IoT devices has significantly complicated AML investigations. As payment networks become more interconnected, there is an increasing need for efficient real-time detection to process large volumes of transaction data \textcolor{black}{on heterogeneous payment systems by different operators such as digital currencies, cryptocurrencies and account-based payments. Most of these mobile payment networks are supported by connected devices, many of which are considered loT devices in the FinTech space that constantly generate data.} Furthermore, the growing complexity and unpredictability of transaction patterns across these networks contribute to a higher incidence of false positives.} While machine learning solutions have the potential to enhance detection efficiency, their application in AML faces unique challenges, such as addressing privacy concerns tied to sensitive financial data and managing the real-world constraint of limited data availability due to data regulations. 
Existing surveys in the AML literature broadly review machine learning approaches for money laundering detection, but they often lack an in-depth exploration of advanced deep learning techniques—an emerging field with significant potential. To address this gap, this paper conducts a comprehensive review of deep learning solutions and the challenges associated with their use in AML. Additionally, we propose a novel framework that applies the least-privilege principle by integrating machine learning techniques, codifying AML red flags, and employing account profiling to provide context for predictions and enable effective fraud detection under limited data availability. Specifically, our approach defines AML-relevant financial profile characteristics and risk indicators to contextualize transactions and assess their associated risks. The proposed context-risk-predict AML (CRP-AML) model demonstrates notable success, achieving an F1 score of 82.51\% on the minority class and nearly doubling the performance of other pattern detection models when the proportion of money laundering records in the dataset drops as low as 0.0005.
\end{abstract}


\section{Introduction}
\label{sec:introduction}

Money laundering is a financial crime that involves concealing financial assets to obscure the illicit origins~\cite{MASCIANDARO1999}. In recent years, combating money laundering has attracted significantly more attention due to the increasing scale and frequency of reported cases. For example, in August 2024, a group of five U.S. nationals was sentenced to imprisonment for laundering over \$8 million. These funds are fraudulently obtained through coordinated cyberattacks on small business computer systems and funnelled into fake companies established across multiple states~\cite{usao2024moneylaundering}.

To combat money laundering, \textcolor{black}{jurisdictions} worldwide have implemented anti-money laundering (AML) laws, mandating large financial institutions such as banks to conduct in-house checks and reporting. In general, AML refers to a collection of policies, laws, and regulations designed to prevent illegally obtained funds from entering the financial system~\cite{TechnologyandAntiMoneyLaundering}. 

\textcolor{black}{The proliferation of mobile payment platforms and smart IoT devices has introduced a significant challenge to AML investigations. \textcolor{black}{As payment networks become increasingly pervasive with smart devices, such as digital payment applications on smartphones, smartwatches, and connected vehicles, vast amounts of transaction data are generated across a wide variety of end devices and platforms~\cite{10506539}.} This massive volume of transaction data presents a pressing need for efficient, real-time AML systems. Furthermore, the growing diversity of payment methods enabled by smart IoT devices and online payment systems, such as peer-to-peer transfers and automated payments, makes it harder to identify patterns indicative of illicit activities. For example, illicit funds can be layered through mobile payment systems on smart devices without the need for criminals to physically visit local banks, thereby hindering local authorities’ ability to investigate the true intention of these transactions.} An illustrative case involves the practice of smurfing, where large sums of illicit money are divided into smaller, seemingly innocent transactions routed through multiple digital wallets across different jurisdictions~\cite{thommandru2023smurfing}. This approach exploits inconsistencies in global AML regulations across jurisdictions to avoid detection. 

Adding to this challenge, criminals often convert illicit funds into virtual assets, such as cryptocurrencies or non-fungible tokens (NFTs), which can be transferred anonymously across platforms and later converted into cash via legitimate exchanges, effectively obscuring their origins~\cite{albrecht2019use}. \textcolor{black}{Traditional AML approaches, which rely heavily on structured and limited datasets, are often ill-equipped to handle the dynamic and decentralized nature of mobile and IoT-enabled transactions.} One example is the 2023 case of Binance~\cite{usdoj2024binance}, where the international cryptocurrency exchange platform pleaded guilty to federal charges of enabling money laundering and sanctions violations, having collected \$1.35 billion in trading fees from these activities.

\begin{figure*}[hb!t]
\begin{center}
    \includegraphics[width=0.95\textwidth]{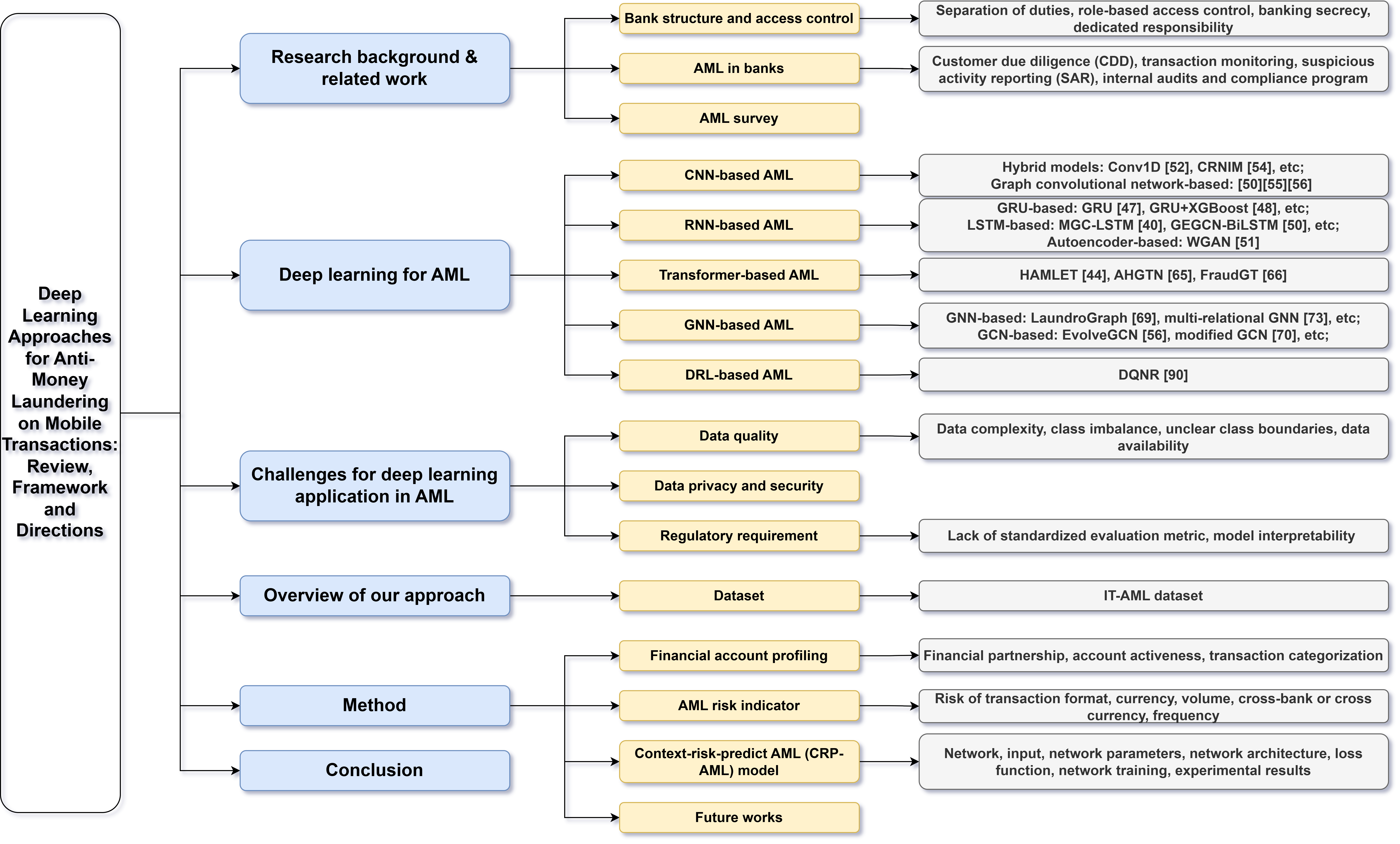}
    \caption{The structure of the survey paper, where we introduce research backgrounds (Section II), review deep learning proposals for AML (Section III), highlight challenges (Section IV), and propose our CRP-AML framework (Sections V-VI).} 
    \label{fig:org}
\end{center}
\vspace{-8mm}
\end{figure*}

In response to the persistent challenge of money laundering, financial institutions have established dedicated AML teams for investigating flagged transactions identified by automated systems that detect suspicious activities daily~\cite{fatf_virtual_assets}. However, conventional rule-based methods have become increasingly inadequate in addressing the growing complexity and volume of digital transactions. These detection mechanisms are often obstructed by inefficiencies and a high rate of false positives, underscoring the urgent need for AML frameworks to integrate advanced technologies.

As financial transactions in the big data era are characterized by their large volume, variety, and velocity, numerous studies have explored innovative advancements in automated AML detection and reasoning techniques. Among the reviewed works, machine learning has consistently emerged as a key tool for enhancing the efficiency and robustness of financial fraud detection, as demonstrated in \cite{TechnologyandAntiMoneyLaundering, ISHIBUCHI20074, westintell, shaikh2018model, 7838276, FARRUGIA2020113318, 10058199}. Furthermore, recent surveys including~\cite{10.1007/s10115-017-1144-z, 9446887, moneylaunderingdetectionusingml, 10.1007/978-3-031-27762-7_8, 10025710}, have focused on machine learning models specifically for AML applications. However, these surveys have provided limited insights into the potential of advanced deep learning techniques.

To address this gap, this study reviews recent advancements in machine learning for AML detection. Specifically, we focus on neural network-based AML methodologies, analyzing the challenges and limitations of current deep learning approaches in detecting and investigating money laundering activities. \textcolor{black}{Moreover, from studying the literature on deep learning for AML, we find that the feasibility of these approaches is heavily compromised by the implicit assumption of readily available global transaction data and the lack of privacy consideration for AML investigation. Therefore, we propose Context-Risk-Predict AML (CRP-AML) framework for implementing the least privilege principle by incorporating machine learning techniques, AML risk indicators and account profiling to provide effective fraud detection under limited data availability.} A detailed comparison of existing surveys and this work is presented in Section \ref{sec: related}.


The main contributions of this paper are summarized as follows:
\begin{itemize}
    \item We examine recent advancements in deep learning approaches for AML, highlighting their potential to outperform traditional machine learning models in both performance and adaptability.
    \item We highlight key challenges for deep learning application in AML in terms of data quality, data privacy and security, and regulatory requirements.
    \item We propose a novel framework comprising three essential components: (1) domain knowledge or statistical insights derived from datasets to account for contextual variations and reduce false positives, (2) customer profiling to refine detection by leveraging behavioral patterns, and (3) advanced deep learning models to enhance efficiency and accuracy.  
\end{itemize}

The structure of this paper is as follows: Section \ref{sec: related} discusses the research background and related work. Section \ref{sec: dlreview} provides a comprehensive review of various deep learning methods for AML. The challenges associated with applying deep learning techniques in AML investigations are analyzed in Section \ref{sec: challenges}. Section \ref{sec: overview} outlines our proposed approach and dataset, while Section \ref{sec: method} presents the details of the proposed methods and experimental results. Finally, the conclusion is presented in Section \ref{sec: conclusion}. The organizational chart of this paper is summarized in Fig. \ref{fig:org}.

\section{Research Background \& Related Work} \label{sec: related}
In this section, we provide an overview of banking structures, access rules and AML practices within financial institutions, drawing on insights from discussions with domain experts at banks in Singapore. This foundational knowledge provides a basis for understanding the role of AML teams and the privacy concerns that arise during AML investigations. Following this, we review related surveys, offering a comparative analysis of their scope and highlighting how our work addresses existing gaps.


\subsection{Bank Structure and Access control}
\begin{figure}[!t]
\begin{center}
    \includegraphics[width=0.40\textwidth]{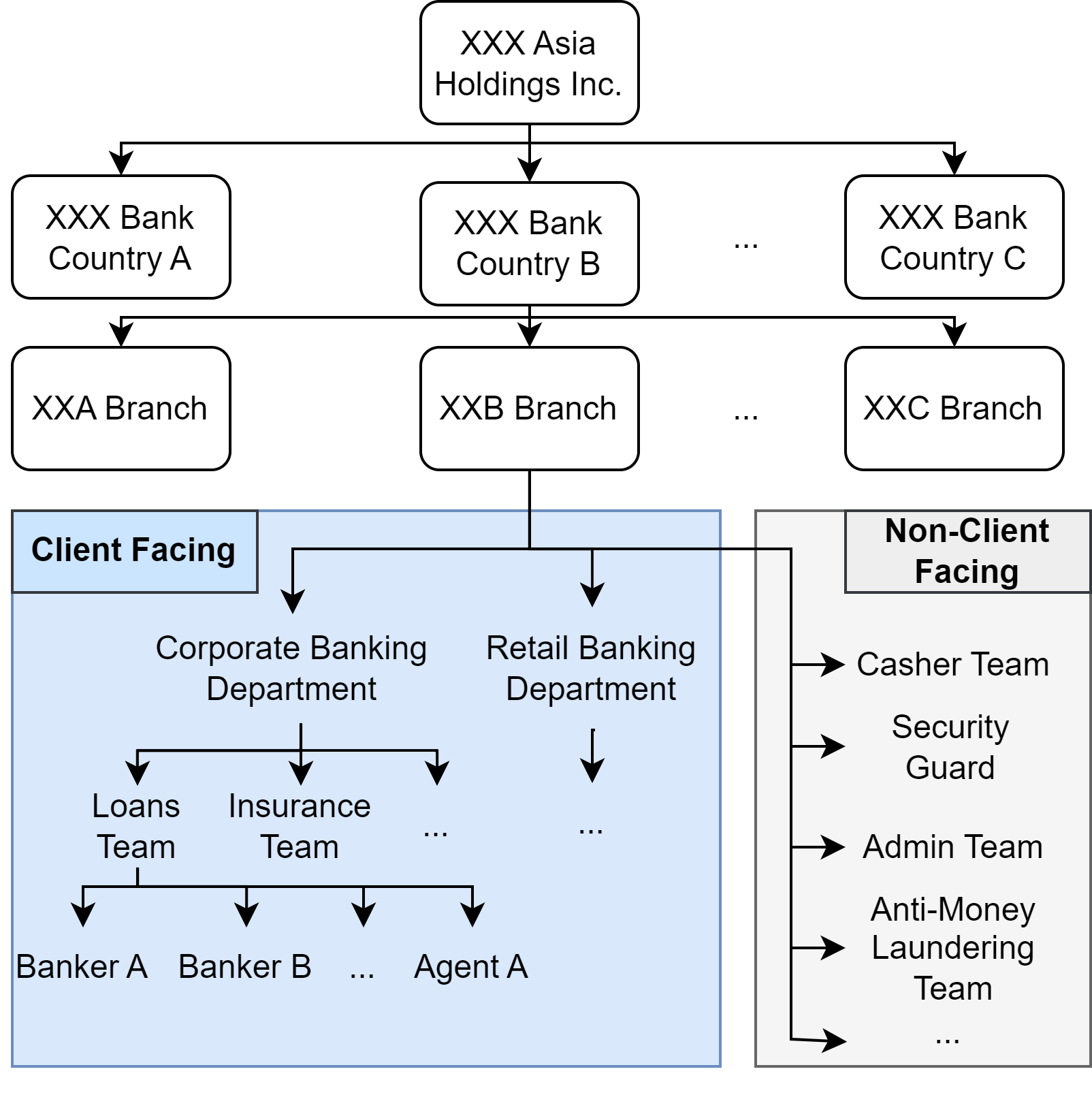}
    \caption{A generalized structure of a typical multinational bank with multiple branches located in different countries~\cite{HSBC}} 
    \label{fig:structure}
\end{center}
\end{figure}

In many nations, financial data are restricted from flowing across countries due to data residency requirements and banking laws imposed by the government, such as the General Data Protection Regulation (GDPR) of the European Union~\cite{GDPR_2022} and the China Personal Information Protection Law (PIPL)~\cite{PIPL_2022}. Hence, most banks separate their banking databases by the country where the account is opened and maintain separate corporate banking departments, retail banking department, and AML team in the country. For clarity, the generalized structure \textcolor{black}{of a typical multinational bank is presented in Fig.~\ref{fig:structure}.}

\textcolor{black}{In addition to traditional physical banking branches, most banks offer digital platforms that allow clients to conduct banking activities through their smart devices. For example, customers can use their smartphones for scan-to-pay functions or access static payment codes on smart wearables. These mobile systems are seamlessly integrated with banking databases through APIs, enabling real-time transaction monitoring across devices and applications. Moreover, smart devices function as both sensors and actuators to provide secure multi-factor authentication and environment sensing, thereby enhancing the security and usability of banking services.}

For the locally held banking database hosting both physical banking sites and digital banking systems, banks generally adopt role-based access control policies centered around two essential guidelines: 
\begin{itemize}
    \item \textbf{{Banking Secrecy Act (BSA)}}: Originating from the Currency and Foreign Transactions Reporting Act of 1970~\cite{bs2022}, similar provisions have been adopted in banking laws worldwide. For example, Section 47 of the Singapore Bank Secrecy Act~\cite{Bankingact_2021} requires bankers holding a Singapore banking license must not disclose client information, including data related to accounts, deposits, investment funds or safe deposit boxes, to unauthorized individuals, except under specific circumstances authorized by the act, such as legal proceedings.
    \item \textbf{The Principle of Least Privilege (PoLP)}~\cite{10.1145/373256.373257}: This minimum access policy grants users access to the minimal privileges necessary while centrally managing and protecting privileged credentials.
\end{itemize}

\textcolor{black}{According to the Banking Secrecy Act \cite{Bankingact_2021}, the standards commonly adopted by multinational banks \cite{HSBC}, and insights from industry experts, banks generally implement the following access control policies to ensure compliance with these guidelines:}
\begin{itemize}
    \item \textcolor{black}{\textbf{Separation of duties}:} As shown in Fig.~\ref{fig:structure}, every bank branch has two different categories of banking departments: client-facing departments that have access to the client data stored in the bank's national database and non-client-facing departments that generally do not have direct access to client data at all.
    \item \textcolor{black}{\textbf{Role-based access control}:} Banks adopt role-based access control, where the job scope defines the access rights of the department, and the head of the team decides the access that the member should have. For example, the head of marketing may have fewer access rights than a normal banker due to differences in jobs and responsibilities.
    \item \textcolor{black}{\textbf{Banking secrecy}:} According to the Banking Secrecy Act~\cite{Bankingact_2021}, banks are not allowed to access customer data that is not within their responsibility or share any private client data they have with anyone who does not have access, violations will have to be reported and justified.
    \item \textcolor{black}{\textbf{Dedicated responsibility}:} The AML team is an independent department that is responsible for investigating suspicious transactions red-flagged by the bank's AML automation rules and has access to all client data for investigation purposes.
\end{itemize}

While bankers in client-facing departments are legally obligated by their banker licenses to adhere to the stringent provisions of the BSA, the AML team enjoys full access to client data for investigative purposes. However, this unrestricted access during the investigation process raises valid concerns about the privacy of sensitive financial information belonging to innocent clients.

\subsection{AML in Banks}
Money laundering represents a critical threat to global financial systems, undermining economic integrity and facilitating various criminal enterprises, including terrorism financing, drug trafficking, and corruption~\cite{MASCIANDARO1999}. The United Nations estimates that approximately 2\% to 5\% of global GDP—equivalent to \$800 billion to \$2 trillion annually—is laundered annually, highlighting the pervasive nature of this issue and the significant challenges it poses to global economic stability~\cite{altman2023realistic}. Despite international efforts to combat money laundering, a substantial proportion of illicit funds continues to evade detection, infiltrating legitimate financial systems and posing ongoing risks to both economic and political stability~\cite{10371347}. 

To address these challenges, governments worldwide have established robust AML regulations and standards, requiring financial institutions to implement comprehensive measures to detect and deter money laundering activities~\cite{muller2007anti}. Central to these frameworks are two primary tasks, i.e., (1) \textit{Suspicion Detection}—identifying transactions that meet predefined risk criteria, and (2) \textit{Offense Determination}—evaluating whether flagged transactions constitute criminal financial activity~\cite{bellomarini2020rule}. These processes form the foundation of AML operations, ensuring that financial systems remain secure and compliant with both local and international regulations. 

To support financial institutions in implementing effective AML strategies, global standards, such as those established by the Financial Action Task Force (FATF)~\cite{fatf_virtual_assets}, along with localized regulatory requirements, offer comprehensive guidance. Key components of recommended AML practices include the following:
\begin{itemize}
    \item \textbf{Customer due diligence}: Financial institutions are required to verify customer identities through know-your-customer (KYC) procedures. This process involves collecting personal identification details, conducting background checks, and assessing risk profiles based on factors, such as geographic location, transaction behavior, and credit history.
    \item \textbf{Transaction monitoring}: Continuous monitoring of customer transactions is mandatory to identify unusual or suspicious patterns. Examples include unusually large cash deposits, small transactions designed to evade reporting thresholds, and transactions linked to high-risk entities or jurisdictions.
    \item \textbf{Suspicious activity reporting (SAR)}: When potential money laundering activities are detected, institutions are obligated to file SARs with local financial intelligence units~\cite{10756563}. These reports alert authorities to suspicious transactions and facilitate further investigation and enforcement.
    \item \textbf{Internal audits and compliance programs}: Financial institutions are required to conduct regular internal audits to ensure their AML programs remain compliant with evolving regulatory standards, effectively mitigate emerging operational deficiencies, and adapt to increasingly sophisticated money laundering techniques~\cite{fatf_virtual_assets}. These audits play a critical role in assessing the robustness of existing systems and identifying potential vulnerabilities, especially in light of the complex laundering schemes enabled by the expanding accessibility and transactional volume of digital and mobile payment platforms.
\end{itemize}

In response to the growing complexities of money laundering, financial institutions, i.e., global banks, have established localized AML teams in each country of operation. These teams are granted extensive data access rights to investigate transactions flagged by automated AML systems while ensuring strict compliance with local regulations, internal policies, and international laws~\cite{amldisclosure, Huang2015}. Typically, the AML process begins with automated systems applying predefined rules to detect suspicious transactions. These flagged transactions are subsequently subjected to detailed review by AML teams to either confirm or dismiss the suspicions. 

While this approach provides a structured framework for combating money laundering, it also faces significant challenges. Specifically, the widespread adoption of mobile payments and increasingly interconnected transaction networks have greatly affected the effectiveness of rule-based systems. Particularly, the increasing complexity of transaction flows makes it difficult to define accurate, static rules, resulting in high false-positive rates in money laundering detection. This not only burdens AML teams with unnecessary investigative work but also allows a significant proportion of criminal activities to slip through unnoticed. The inefficiencies highlight the urgent need for more advanced detection mechanisms.

\begin{figure}[hb!t]
\begin{center}
    \includegraphics[width=0.45\textwidth]{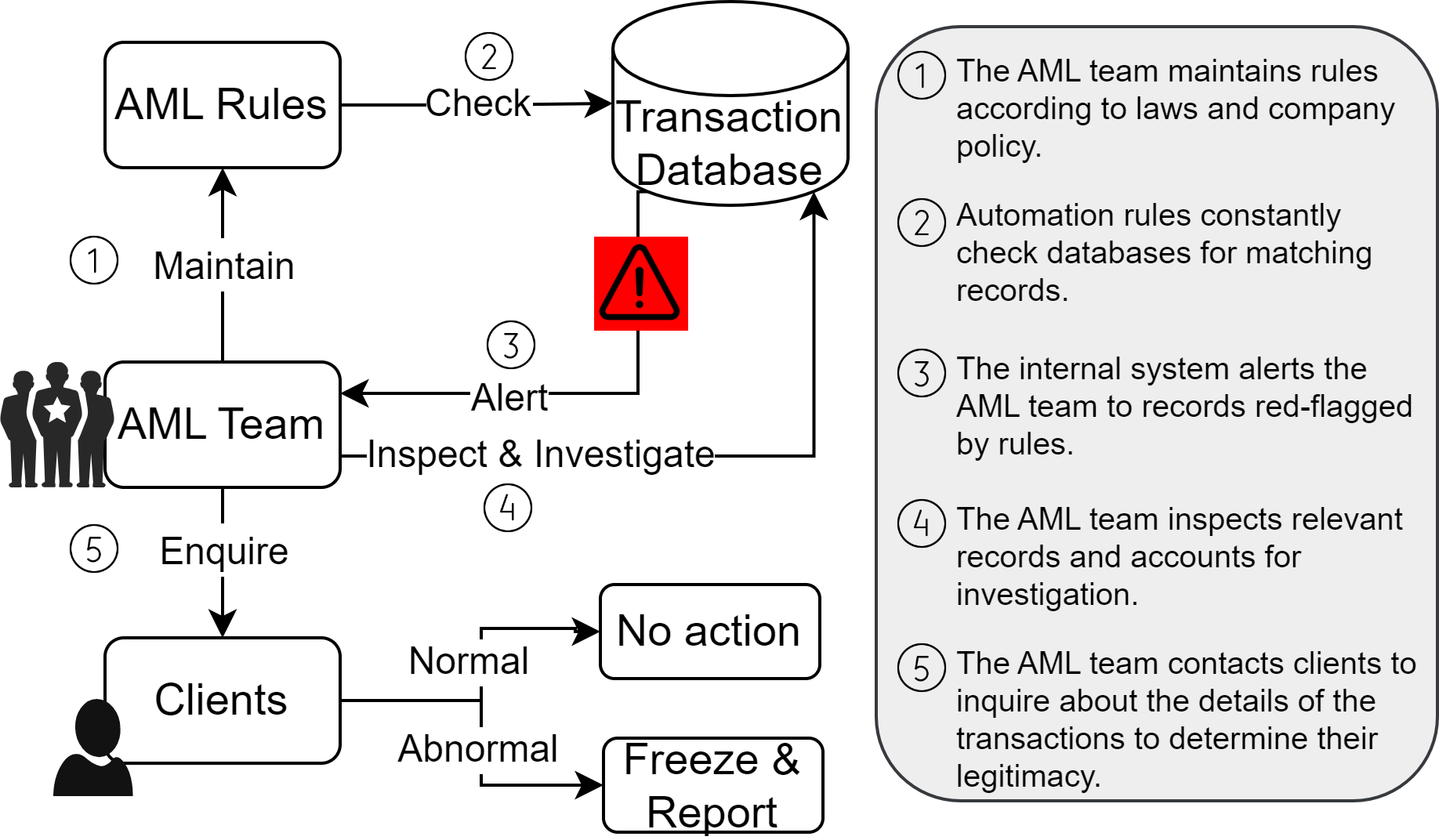}
    \caption{\textcolor{black}{An illustration of the general AML investigation flow in banks, starting from daily maintenance and reporting to reviewing cases flagged by AML systems.}}
    \label{fig:amlflow}
\end{center}
\end{figure}

\begin{table*}[!t]
    \centering
    \caption{A comparison of related surveys and this work}
    \begin{tabular}{|m{1.75cm}|m{0.75cm}|m{0.75cm}|m{2.5cm}|m{10cm}|}
        \hline
        \textbf{Scope} & \textbf{Ref.} &  \textbf{Year} & \textbf{Emphasis} & \textbf{Summary} \\ \hline
        Machine learning approaches for AML &\cite{10.1007/s10115-017-1144-z} & 2018 & Integrating machine learning into AML processes across various aspects of KYC guidelines & A comprehensive survey categorizes machine learning-based AML approaches across various aspects of KYC guidelines. It thoroughly documents the strengths and limitations of different machine learning algorithms in AML and provides a clear explanation of the AML process, from data preprocessing to prediction. However, as this survey was published in 2018, it does not cover the significant advancements made in the field of machine learning since its publication. Nevertheless, it serves as an excellent introductory resource for researchers new to the field of machine learning-based AML. \\ \hline
        Deep learning approaches for AML and XAI techniques &\cite{9446887} & 2021 & Enhancing the explainability of deep learning-based AML models &  This survey focuses on enhancing explainability in deep learning-based AML approaches, emphasizing how various explainable AI (XAI) techniques can be integrated into deep learning models to improve interpretability. It also examines the challenges associated with detecting money laundering. While the survey effectively highlights the critical issue of explainability, which is essential for the broader adoption of machine learning in AML, it lacks examples of proposals that have successfully applied XAI techniques alongside machine learning for AML detection.\\ \hline
        Machine learning approaches for AML &\cite{moneylaunderingdetectionusingml} & 2022 & Comparing the performance of machine learning algorithms on cryptocurrency dataset & This study evaluates the performance of various machine learning methods on the Elliptic dataset. While it offers a performance comparison of baseline models such as GCN, RF, and XGBoost, it does not explore more advanced machine learning models built upon these base models, which tend to achieve superior performance in AML tasks.\\ \hline
        Machine learning approaches for detecting financial fraud &\cite{10.1016/j.eswa.2021.116429} &  2022 & Anomaly detection for detecting financial frauds & This comprehensive survey examines anomaly detection techniques for financial fraud, covering developments from 2002 to 2020. While it thoroughly documents various types of financial fraud and the machine learning approaches used to address them, its primary focus is on financial fraud in general, with limited elaboration on proposals specifically tailored to AML.\\ \hline
        Machine learning approaches for AML &\cite{10.1007/978-3-031-27762-7_8} & 2023 & Supervised and unsupervised machine learning techniques for AML & This broad survey classifies machine learning-based AML methods into supervised and unsupervised approaches. It examines a variety of techniques, ranging from traditional machine learning models such as decision trees, SVMs, and logistic regression to deep learning methods like GCNs and CNNs. However, the survey lacks a specific focus on deep learning-based approaches. \\ \hline
        Machine learning approaches for AML &\cite{10025710} & 2023 & Supervised and unsupervised machine learning techniques for AML, feasibility challenges & This technical survey reviews both supervised and unsupervised machine learning solutions for client and transaction AML monitoring. It also explores the challenges associated with applying machine learning to AML detection and discusses various evaluation metrics suitable for AML tasks. However, the survey does not specifically focus on deep learning-based solutions.\\ \hline
        Deep learning approaches for detecting credit card fraud &\cite{10595068} & 2024 & Deep learning models for detecting credit card fraud &  A comprehensive survey on deep learning methods for credit card fraud detection, presenting the theoretical foundations of various deep learning techniques. It highlights critical challenges and provides a comparative analysis of different approaches using a European credit card dataset. However, its primary focus is on credit card fraud detection, rather than AML tasks.\\ \hline
        Deep learning approaches for AML &This work & - & Deep learning models for AML: challenges and a potential framework to address them &  We review recent advancements in deep learning for AML detection, highlighting key challenges for improving detection performance: achieving a low false positive rate with high detection accuracy. To address these challenges, we propose a framework integrating domain knowledge, customer profiling, and advanced deep learning models to enhance efficiency and reduce false positives.\\ \hline
    \end{tabular}
    \label{tab:surveycomparison}
\end{table*}

Machine learning-based models offer a promising solution by enhancing detection accuracy, adapting to evolving transaction patterns, and reducing the operational workload on AML teams. Such models can mitigate the limitations of rule-based systems, ultimately strengthening the overall effectiveness of AML operations. \textcolor{black}{However, the feasibility of machine learning-based models is heavily compromised by the implicit assumption of readily available global transaction data and the lack of privacy consideration for AML investigation. To illustrate, as depicted in Fig.~\ref{fig:amlflow}, existing machine learning-based solutions primarily contribute to the suspicion assessment phase of AML in steps 2-3, constructing a model on the transaction database to detect global fraud patterns, with the assumption that all transaction flows are traceable and recorded in the same database. However, in reality, banks are restricted by banking laws to possess only the local transaction data entrusted to them by their clients. This makes it nearly impossible to obtain transaction data across multiple banks and geographies, thus preventing a comprehensive view of all financial asset flows. Hence, this restricted view of data impedes the feasibility of such models in their application for real-life AML practices.
Furthermore, these solutions do not address the potential privacy risks associated with exposing confidential financial data to third parties, such as the AML team, during fraud determination and verification in steps 4-5. Any ill-intentioned individual on the team could potentially manipulate sensitive data records or infer confidential information about other clients given their privileges. Hence, a novel machine learning approach is required to balance efficiency with the stringent privacy protections necessary for handling sensitive financial data in real-world settings.}

\subsection{AML surveys}

Many existing surveys on AML detection are limited in scope or a lack of technical details. Some offer broad, qualitative reviews that categorize studies by publication volume or type without providing much technical insight. For example, \cite{goecks} compiles AML and financial fraud detection research from 2007 to 2021, including qualitative, quantitative, and mixed method studies, but does not go into specific methodologies. Similarly, \cite{Soltani} analyzes trends in fraud detection research using keyword searches and general categorizations, providing little actionable detail. Other surveys emphasize broader financial applications and often overlook AML-specific challenges. For example, \cite{OZBAYOGLU2020106384} classifies research in financial domains such as algorithmic trading, risk assessment, fraud detection, and portfolio management. Although it identifies trends and the applicability of models, it lacks a focused discussion on AML detection. Similarly, \cite{10.1016/j.eswa.2021.116429} reviews anomaly detection methods in financial fraud but focuses primarily on credit card and auto insurance fraud, with minimal attention to AML-specific issues \textcolor{black}{such as compliance with data privacy laws and regulatory requirements}.

Fortunately, recent advancements in machine learning have led to a growing number of studies focusing on the AML domain, where machine learning has emerged as a pivotal solution for addressing inefficiencies and enhancing the accuracy of AML procedures. Thus, numerous surveys have investigated the progress in applying machine learning techniques to AML processes. For example, \cite{10.1007/s10115-017-1144-z} provides a comprehensive overview of the strengths and limitations of various machine learning algorithms applied to AML tasks. The authors also discuss preprocessing steps and examine a range of approaches, including link analysis, behavior modeling, anomaly detection, and more. Furthermore, the survey highlights a critical limitation, i.e., the effectiveness of machine learning algorithms is highly dependent on the quality of the available data. 

As of 2018, the scarcity of real-world datasets and confirmed case labels posed a considerable challenge to the development and evaluation of AML models~\cite{10371347}. Much of the AML research relied on undisclosed banking datasets or unverifiable labels, making it difficult to establish a standardized comparison of model performance. \textcolor{black}{Furthermore, AML datasets often suffer from severe class imbalance, which can lead to an overemphasis on the majority of benign transactions~\cite{10477569}. Additionally, the classification boundaries for transaction records are often unclear, as suspect accounts may conceal a few laundering activities within thousands of legitimate transactions through layering strategies.} This lack of high-quality data emerged as a critical bottleneck in advancing robust and automated AML solutions.

Recent advancements in the availability of standardized open AML datasets, such as the Elliptic dataset~\cite{elliptic_dataset} and the IBM Transactions for AML (IT-AML) dataset~\cite{altman2023realistic}, have significantly improved the landscape for training financial transaction-related models. For example, \cite{moneylaunderingdetectionusingml} evaluates the performance of machine learning techniques, including deep neural networks (DNN), random forest (RF), k-nearest neighbors (KNN), and naive bayes (NB), on the Elliptic dataset. Meanwhile, the works of \cite{10.1007/978-3-031-27762-7_8} and \cite{10025710} explore general challenges in AML and provide broad classifications of AML monitoring techniques, focusing on the distinctions between supervised and unsupervised learning approaches. In contrast, \cite{10595068} and \cite{9446887} adopt narrower scopes in their surveys, in which the former examines the specific application of machine learning to credit card fraud detection and the latter investigates the explainability of AML models.

In comparison, this study focuses on recent advancements in deep learning approaches for AML detection, a promising area within machine learning that has shown superior performance over traditional techniques in terms of adaptability and accuracy across various domains~\cite{10.1007/s10115-017-1144-z}. From studying the literature on deep learning for AML, we found that the feasibility of these approaches is heavily compromised by challenges such as data quality, privacy, security, and regulatory requirements. For example, the implicit assumption of readily available global transaction data and the lack of privacy consideration for AML investigation are some of the major real-world concerns in applying these models. To address these issues, we propose a novel framework comprising three essential components: 
\begin{enumerate}
    \item \textbf{Domain Knowledge and Statistical Insights}: Leveraging contextual variations in datasets to reduce false positives.
    \item \textbf{Customer Profiling}: Refining detection by incorporating behavioral patterns.
    \item \textbf{Advanced Deep Learning Models}: Enhancing efficiency, scalability, and accuracy.
\end{enumerate}
By doing so, this integrated framework aims to overcome the limitations of existing methods, offering a pathway toward more effective and robust AML systems. A comparison of related surveys and this work is presented in Table \ref{tab:surveycomparison}.

\section{Deep learning for AML} \label{sec: dlreview}
Machine learning has become a widely adopted strategy for AML due to its effectiveness in detecting fraud within large transaction datasets. Various machine learning techniques have been explored in the AML domain, including fuzzy genetics-based machine learning (GBML) algorithms~\cite{ISHIBUCHI20074}, support vector machines (SVM)~\cite{westintell}, social network analysis~\cite{shaikh2018model}, behavior modeling~\cite{TechnologyandAntiMoneyLaundering}, and deep learning~\cite{7838276}. However, traditional machine learning methods exhibit significant limitations that hinder their performance in AML tasks. For example, fuzzy models struggle with the nonlinear complexities of AML problems, while genetic algorithms (GA) encounter challenges such as parameter tuning, coding complexity, and slow convergence. Similarly, despite their robustness and generalization capabilities, SVM models become computationally infeasible when applied to the high-dimensional datasets common in AML tasks~\cite{xia2022novel}.

Meanwhile, deep learning has emerged as a transformative tool in AML due to its ability to overcome many of the limitations associated with traditional machine learning methods~\cite{10570372}. By automatically extracting high-dimensional and complex features, deep learning models effectively identify complex patterns within large, dynamic datasets. Its core strength lies in neural network architectures composed of interconnected layers of artificial neurons, optimized through advanced algorithms such as backpropagation and stochastic gradient descent. These networks are capable of learning multi-dimensional representations of data, capturing complex relationships, and extracting subtle features without the need for extensive feature engineering.

In the context of AML, deep learning models are particularly effective due to their ability to process complex data structures and identify anomalous patterns in transactional data. For example, attention-based recurrent neural network (RNN) frameworks in natural language processing have been utilized to analyze AML transactions, capturing both lexical contexts and sentence-level information. These models have significantly reduced the workload of AML investigations by automating preliminary analyses~\cite{Han2018NextGenAD}. 

Key models in deep learning include:
\begin{itemize}
    \item \textbf{Convolutional neural network (CNN):} CNNs are highly effective architectures for capturing spatial hierarchies, utilizing convolution and pooling layers to extract meaningful patterns from data. Through averaging or downsampling, CNNs iteratively optimize feature maps by training updates, enabling the encapsulation of crucial spatial features~\cite{9490350}. This capacity to automatically learn and refine hierarchical spatial representations makes CNNs particularly well-suited for tasks such as image recognition, object detection, and image segmentation. 
    \item \textbf{RNN:} The strength of RNNs lies in their ability to model dependencies in sequential data, with hidden layers that process input sequences in a temporal order. This capability makes RNNs particularly well-suited for tasks such as time-series analysis and natural language processing, where the meaning or value of each input is influenced by the context established by both preceding and subsequent inputs~\cite{9739684}. Popular variants of RNNs, such as Long Short-Term Memory (LSTM) networks and Gated Recurrent Units (GRUs), have been developed to overcome limitations such as vanishing gradients, enabling more efficient learning of long-range dependencies within sequences.
    \item \textbf{Transformers:} Unlike RNNs, which process input in a sequential manner, transformers utilize self-attention mechanisms to capture dependencies among input units simultaneously, irrespective of their positions within the sequence~\cite{tatulli2023hamlet, 10679152}. This parallel processing not only enhances efficiency but also enables transformers to capture longer-range dependencies more effectively than RNN-based models. By incorporating techniques such as multi-head attention and positional encoding, transformer-based models have achieved state-of-the-art performance in tasks such as machine translation and question answering~\cite{10670196}.
    \item \textbf{Graph neural networks (GNN):} GNNs are specifically designed to process graph-structured data, which consist of nodes and edges that interconnect them~\cite{10477569}. By leveraging the inherent relationships between connected entities, GNNs are capable of capturing interdependencies and patterns through message-passing mechanisms, which learn both node features and the relational context of input nodes. This ability to model information flow within a network makes GNNs particularly effective for tasks such as drug discovery, fraud detection, and traffic prediction.
    \item \textbf{Deep reinforcement learning (DRL) models:} DRL models combine reinforcement learning with deep learning techniques to address complex decision-making problems and manage high-dimensional data~\cite{9272624}. Using deep neural networks to learn the mapping between observed environmental states and optimal actions over time, DRL models are able to generalize to complex scenarios, such as game play and autonomous driving.
\end{itemize}

While each of these techniques offers distinct advantages across various application domains, most money laundering datasets are organized as time-series transaction data, categorized by the banking accounts associated with the transactions~\cite{mltechnique}. This structure provides RNNs and transformers with a natural advantage, as they are particularly suitable at processing sequential data and capturing long-range dependencies between accounts. However, GNN-based methods have also gained significant traction in AML research, focusing on identifying abnormal money flows or transaction patterns that deviate from conventional financial behavior. By leveraging the relational context between entities, GNNs provide a complementary perspective to sequential modeling approaches, making them an increasingly important tool in the fight against money laundering. In the following subsections, we explore the applications of RNNs, transformers, and GNNs in AML tasks, highlighting their strengths, limitations, and real-world implementations.

\begin{table*}[!t]
    \begin{center}
    \caption{A comparison of the RNN-based AML proposals}
    \begin{tabular}{|m{0.4cm}|m{1.35cm}|m{1.cm}|m{1.85cm}|m{2.5cm}|m{8.25cm}|}
        \hline
        \textbf{Ref.} &  \textbf{Dataset} & \textbf{Dataset size} & \textbf{Best model} & \textbf{Evaluation metrics} & \textbf{Insights} \\ \hline
        \cite{10814854} & Private & 138,000 & GRU & AUC-ROCs & SimpleRNN networks are prone to the vanishing gradient problem, whereas GRU outperforms both SimpleRNN and LSTM models on the anonymous small-scale Brazilian AML dataset. \\ \hline
        \cite{10724384} & IT-AML (Low-illicit medium size subset) & 31,251,483 & GRU+ XGBoost & Accuracy, precision, recall, and F1-score & SMOTE can effectively address class imbalance, while hybrid models consistently outperform standalone deep learning models.\\ \hline
        \cite{xia2022novel} & Elliptic & 234,355 & MGC-LSTM & Precision, recall, F1, micro-precision, micro-recall, micro-F1 & The integration of GCN and LSTM models enables the capture of both spatial and temporal dependencies, resulting in more accurate money laundering predictions. \\ \hline
        \cite{sym16030378} & Elliptic; OGB-Arxiv & 234,355; 1,166,243 & MDGC-LSTM & Micro-precision, micro-recall, micro-F1, macro-precision, macro-recall, macro-F1  &  By leveraging dynamic-GCN, the MDGC-LSTM model is better equipped to handle AML prediction tasks involving dynamically evolving laundering patterns. \\ \hline
        \cite{financialfraudtransaction} & Elliptic; Credit card fraud (CCF) & 234,355; 284,807 & GEGCN-BiLSTM & Precision, recall, F1, micro-precision, micro-recall, micro-F1 &  By utilizing BiLSTM to capture temporal dependencies from both preceding and subsequent timestamps, while employing GEGCN to extract global spatial contextual relationships between transactions, the GEGCN-BiLSTM model outperforms models such as GEGCN, GCN-LSTM, and CNN\\ \hline
        \cite{9446893} & Private & 4889 & AE + VAE + WGAN & Accuracy, precision, recall, F1, AUC, FPR &  Autoencoders can be effective in reducing false positives, while WGANs are useful for addressing data imbalance. Additionally, optimizing the anomaly score threshold can help maximize recall while maintaining high precision.\\ \hline
    \end{tabular}
    \label{tab:RNN}
    \end{center}
\end{table*}

\subsection{CNN-based AML}

\begin{figure}[hb!t]
\begin{center}
    \includegraphics[width=0.45\textwidth]{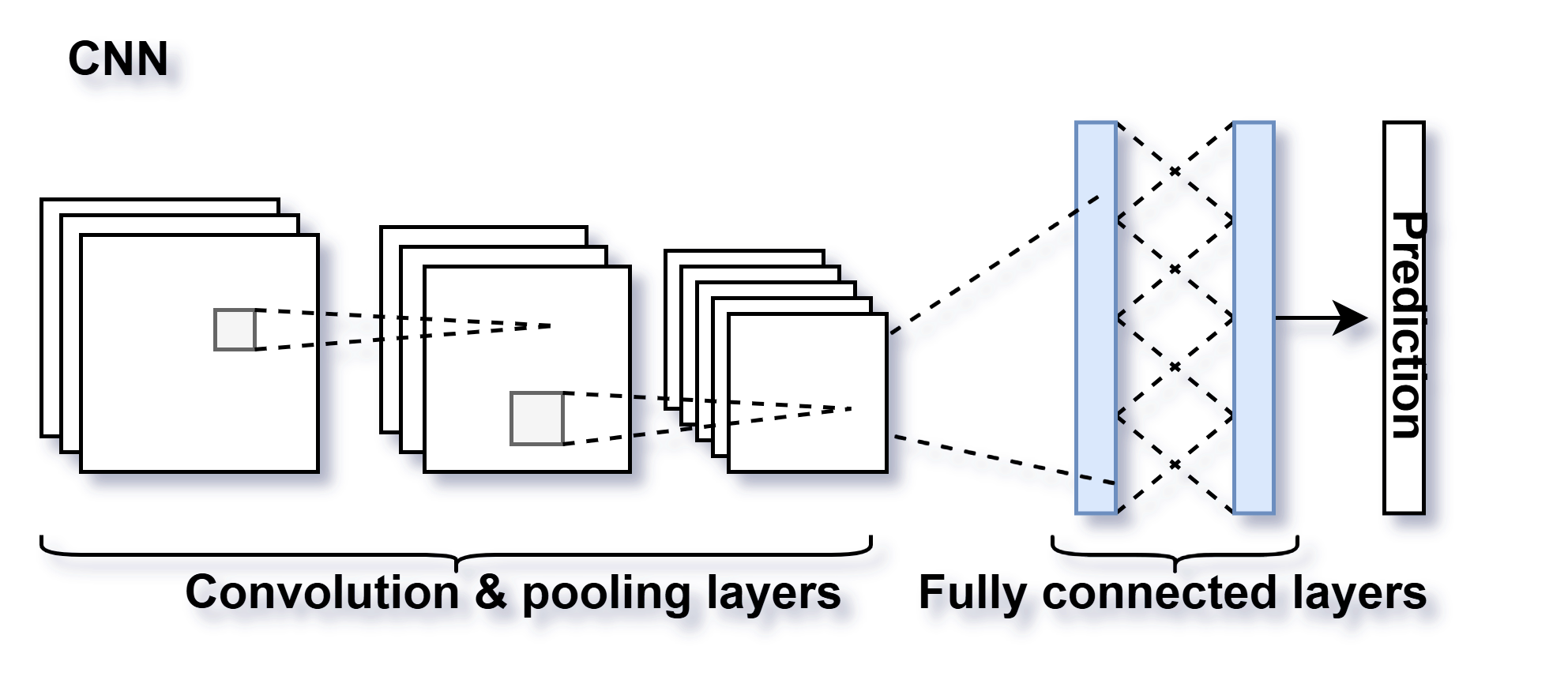}
    \caption{\textcolor{black}{An illustration of typical CNN architecture comprises convolutional and pooling layers, followed by fully connected dense layers for generating predictions.}}
    \label{fig:cnn}
\end{center}
\vspace{-5mm}
\end{figure}

Although originally designed for image processing, CNN-based methods have demonstrated considerable effectiveness in detecting money laundering activities, owing to their ability to efficiently extract features and recognize complex patterns. \textcolor{black}{As shown in Fig. \ref{fig:cnn}}, by transforming financial data into structured formats such as transaction matrices or heatmaps, CNNs can uncover spatial and temporal correlations indicative of illicit activities. Furthermore, their capability to automatically learn critical features from high-dimensional data makes them particularly suitable for analyzing the vast transaction datasets generated by financial institutions. 

For example, Kute et al.\cite{kute2024explainable} proposes a Conv1D CNN classifier combined with the model-agnostic explainability technique SHAP\cite{lundberg2017unifiedapproachinterpretingmodel} to effectively detect money laundering while also enhancing the interpretability of predictions. However, although the Conv1D CNN model achieved an overall F1 score of 0.7823, the evaluation is conducted on a synthetically generated dataset with fewer than 30,000 records. As a result, its applicability to real-world scenarios remains uncertain, particularly in handling imbalanced datasets and more complex transaction patterns typical of large-scale financial systems. 

In another study, \cite{yu2024deeplearningcrossbordertransaction} design and experiment various CNN architectures, including CNN, Deep CNN, CNN-LSTM, and hybrid CNN-GRU, and proposes a hybrid Convolutional-recurrent neural integration model (CRNIM) to optimize AML rules for enhanced money laundering detection performance. By employing an unsupervised contrastive learning approach to automatically extract features and dynamically optimize rules based on cluster analysis of recent transactions, the AML system demonstrated the ability to adapt effectively to emerging laundering techniques. Experimental results on the Elliptic dataset revealed that as model complexity increased, both detection accuracy and the area under the curve (AUROC) of true positives plotted against false positives improved. The proposed CRNIM outperformed all tested CNN variants, achieving 97.1\% overall accuracy and an AUROC of 0.94. However, the evaluation metrics could benefit from incorporating minority F1 scores to assess performance on rare but critical money laundering cases.

Furthermore, when combined with graph-based representations, CNNs can analyze the complex relationships between accounts and transactions, providing a comprehensive understanding of the events that occur. A common approach that leverages the strengths of convolution and graph structures for AML classification tasks is graph convolutional networks (GCNs), i.e., \cite{du2018topologyadaptivegraphconvolutional, pareja2020evolvegcn, financialfraudtransaction}, which will be discussed in detail in subsection \ref{subsec: GNN}.

\subsection{RNN-based AML} 

\begin{figure}[hb!t]
\begin{center}
    \includegraphics[width=0.35\textwidth]{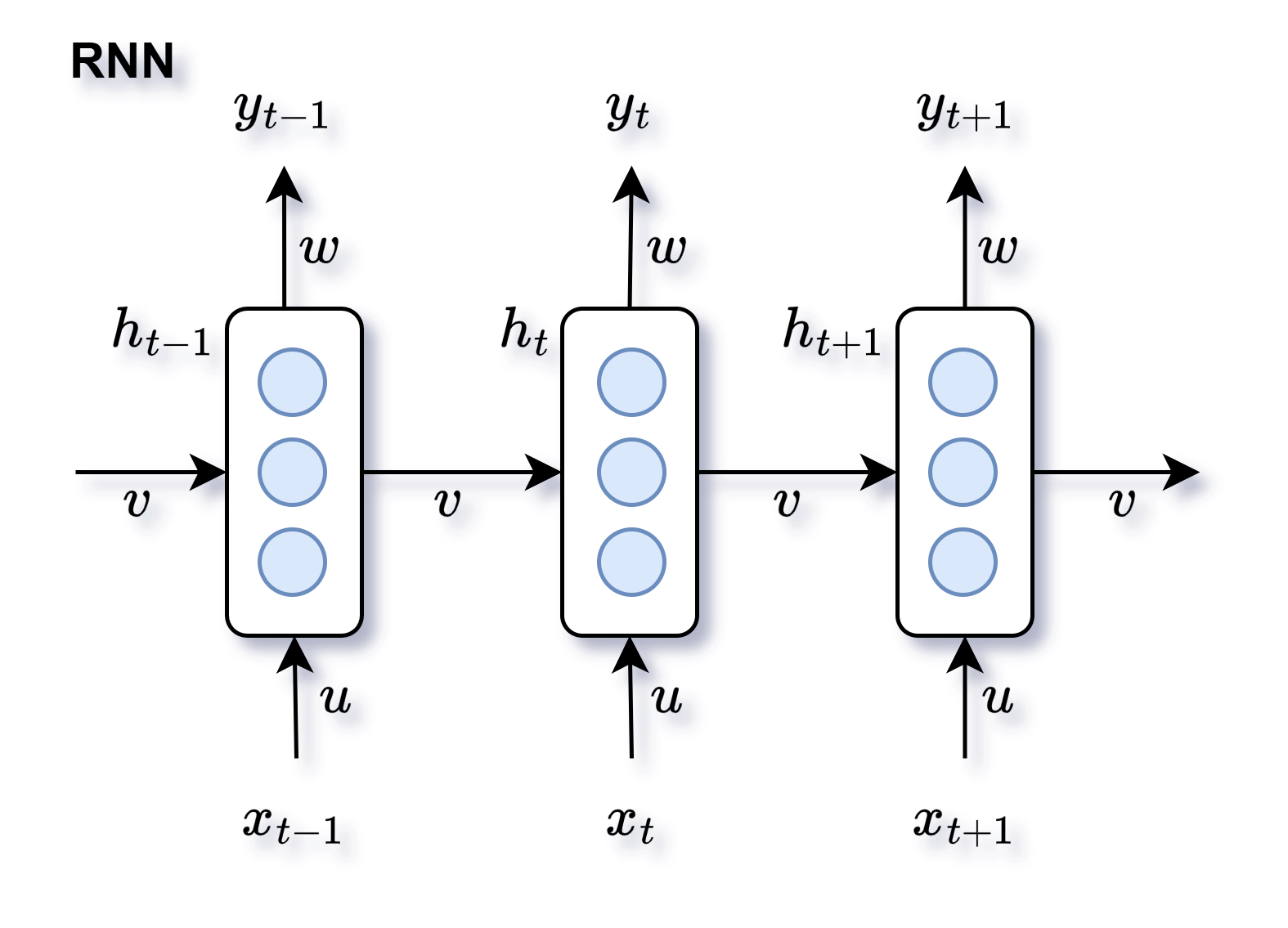}
    \caption{\textcolor{black}{An illustration of the RNN sequential processing at a specific timestamp, where \( h \) represents the hidden state, \( w \), \( v \), and \( u \) denote the shared weight matrices, \( x_t \) is the input vector for each element in the input sequence, and \( y_t \) is the corresponding prediction.}}
    \label{fig:rnn}
\end{center}
\vspace{-5mm}
\end{figure}

RNN-based methods are particularly well-suited for detecting money laundering activities due to their ability to model dependencies in sequential and temporal data, \textcolor{black}{as shown in Fig. \ref{fig:rnn}}. By processing input sequences in an ordered manner through hidden layers, RNNs are highly effective for time-series analysis, where the meaning of each transaction depends on the context established by preceding and subsequent events~\cite{tatulli2023hamlet}. This characteristic is crucial in analyzing financial transaction patterns, as money laundering schemes often involve evolving behaviors like structuring transactions or layering funds across multiple accounts over time. Popular variants of RNNs, such as long short-term memory (LSTM)~\cite{10.1162/neco.1997.9.8.1735} networks and gated recurrent units (GRUs)~\cite{cho2014learningphraserepresentationsusing}, address challenges such as vanishing gradients, further improving the learning of long-range dependencies within transaction sequences. These capabilities allow RNNs to detect anomalies and deviations from normal transaction patterns, which are often indicative of illicit activities. When integrated with graph-based methods, RNN can simultaneously analyze temporal and relational data, offering a robust framework for identifying complex laundering schemes and adapting to evolving money laundering techniques in dynamic financial networks.

For example, \cite{10814854} examines the use of RNN-based models, including SimpleRNN, LSTM, and GRU, to identify money laundering typologies on an anonymized dataset provided by the Civil Police of the State of Pernambuco, Brazil. The study highlights that SimpleRNN models face significant challenges in processing long input sequences due to vanishing gradient issues, whereas LSTM and GRU models demonstrate superior performance. Among these, GRU achieves the highest results, consistently maintaining an AUC-ROC score above 0.7. However, the dataset employed in the study is both confidential and limited in scale, comprising only around 138,000 transaction records from fewer than 200 accounts. This constraint undermines the generalizability and broader applicability of the findings for evaluating anti-money laundering models.

In another study, \cite{10724384} investigates the use of various RNN models for money laundering detection on the IT-AML dataset~\cite{altman2023realistic}. To address class imbalance, the authors applied the synthetic minority oversampling technique (SMOTE)~\cite{Chawla2002} and evaluated both standalone models, such as LSTM, GRU, and BiLSTM, and hybrid models. The authors conclude that standalone models are inadequate for handling the complexities of AML tasks, as evidenced by their low recall and F1 scores. Among the hybrid models, the combination of GRU's deep learning capabilities and XGBoost's~\cite{Chen2016} robust feature engineering achieved the highest F1 score.

Furthermore, the authors in \cite{xia2022novel} highlight that the graph structure and spatial dependencies inherent in money laundering data can be effectively modeled using graph convolutional networks (GCN). They propose a novel prediction model, MGC-LSTM, which integrates GCNs with LSTM networks to capture both spatial and temporal dependencies associated with money laundering activities. Experiments on the Elliptic dataset demonstrate that MGC-LSTM outperforms traditional machine learning methods, as well as models such as LSTM, GCN, GRU, GCN-GRU, and EvolveGCN~\cite{pareja2020evolvegcn}, in terms of precision, recall, and F1 score. However, the proposed model primarily relies on static money laundering transaction graphs for graph convolution, which results in issues such as inefficient utilization of dynamic temporal features and information leakage when applied to prediction tasks on dynamic time series datasets~\cite{sym16030378}. 

Hence, the authors in \cite{sym16030378} build upon the techniques introduced in \cite{10.1145/3534678.3539300} and \cite{xia2022novel} by incorporating time attributes into graphs, transforming them into timestamped dynamic graphs, and proposing a dynamic variant of MGC-LSTM, termed MDGC-LSTM, which outperforms the original MGC-LSTM. Similarly, in \cite{10.1007/s11063-022-10904-8}, the authors proposed a temporal-GCN for the classification of illicit transactions in the Elliptic dataset, which combines LSTM with TAGCN~\cite{du2018topologyadaptivegraphconvolutional}, a variant of the GCN model. This model demonstrates superior performance over both Skip-GCN and EvolveGCN~\cite{pareja2020evolvegcn}. Nonetheless, the models in \cite{10724384, xia2022novel, sym16030378, 10.1007/s11063-022-10904-8} rely on evaluation metrics that assess overall performance across both illicit and licit data, which limits their effectiveness in addressing the issue of false positives—a significant challenge for many AML systems.

At the same time, a recent study presented in \cite{financialfraudtransaction} proposes the GEGCN-BiLSTM model, which integrates global enhanced graph convolution network (GEGCN) with BiLSTM for predictive analysis. This model utilizes BiLSTM to capture temporal dependencies from both preceding and subsequent timestamps while employing GEGCN to extract global spatial contextual relationships between transactions. Experimental results on the Elliptic dataset demonstrate that GEGCN-BiLSTM outperforms models such as GEGCN, GCN-LSTM and CNN. However, like the approaches mentioned above, it relies on evaluation metrics that assess overall dataset performance rather than focusing on the minority class, thereby limiting its applicability in addressing the challenges posed by imbalanced data.

An alternative RNN-based approach for AML tasks is proposed in \cite{9446893}, where the authors utilize autoencoders (AE) and variational autoencoders (VAE) and significantly reduce the false positive rate by half. To address data imbalance, the method employs a wasserstein generative adversarial network (WGAN)~\cite{arjovsky2017wassersteingan} to generate synthetic fraudulent transactions. Additionally, the authors introduce an anomaly score threshold, referred to as the Recall-First Threshold (RFT), for maximizing recall score for all fraudulent transactions while achieving the highest possible precision, thereby mitigating false positives. To validate the model performance, experiments were conducted on a subset of 4,889 transaction records derived from a collaborative project between the University of Nottingham (Malaysia campus) and a local Malaysian bank. While the approach provides a promising avenue for incorporating RNNs into AML, its overall effectiveness is constrained by the limited accessibility of the dataset, the small sample size, and the static nature of the data used in the evaluation. 

To summarize, Table \ref{tab:RNN} presents a comparison of RNN-based models along with key insights derived from these proposals.

\subsection{Transformers-based AML}

\begin{figure}[hb!t]
\begin{center}
    \includegraphics[width=0.45\textwidth]{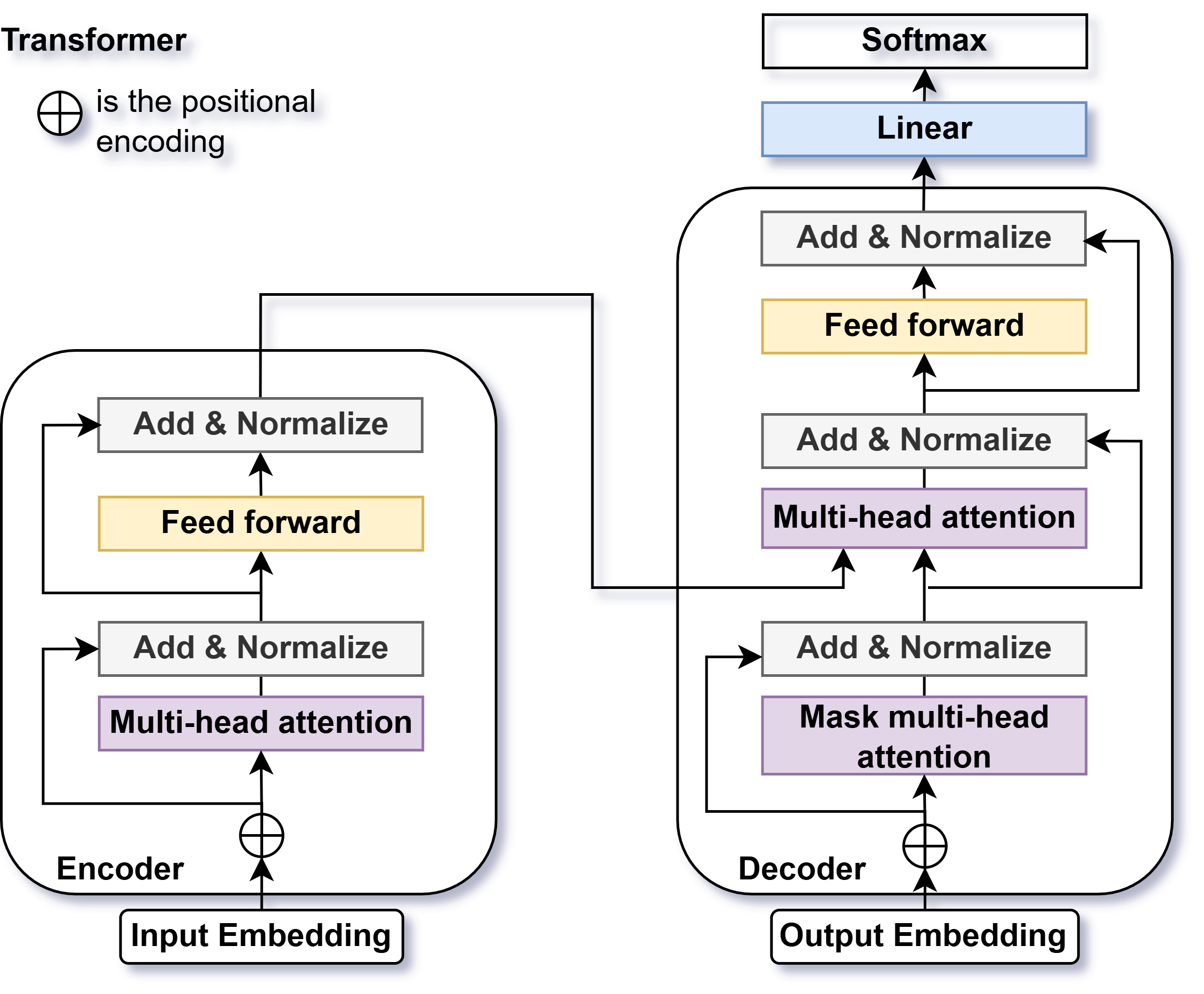}
    \caption{\textcolor{black}{An illustration of a typical Transformer architecture comprises encoder and decoder components for making predictions.}}
    \label{fig:transformer}
\end{center}
\vspace{-5mm}
\end{figure}

Transformers have recently emerged as a groundbreaking architecture in deep learning, particularly for their ability to model complex relationships, capture long-range dependencies, and process large-scale data efficiently~\cite{han2022survey}. Unlike traditional sequential models like RNNs, transformers rely on self-attention mechanisms, \textcolor{black}{as shown in Fig. \ref{fig:transformer},} allowing them to focus on critical relationships within the data, regardless of their position in a sequence. This makes transformers particularly suited for analyzing transactional data, capturing both global and local dependencies and providing a comprehensive understanding of the data~\cite{10531073}. This is essential in money laundering detection, where complex schemes such as layering, circular transactions, or cross-border activities often involve intricate and non-linear relationships. Additionally, transformers can process transaction sequences in parallel, significantly improving computational efficiency and scalability for large datasets typically generated by financial institutions.

Recent transformed-based proposals for AML includes~\cite{tatulli2023hamlet, xu2023illegal, lin2024fraudgt}. In \cite{tatulli2023hamlet}, the authors proposed a scalable deep learning model, HAMLET, which employs a hierarchical transformer with an attention mechanism at both the transaction and sequence levels. By encoding transaction features, mapping them to a predefined vocabulary, and subsequently re-encoding them at the sequence level, HAMLET effectively captures complex relationships between transaction events. To validate, the model was evaluated on a synthetic dataset provided by collaborators~\cite{9758694}, simulating clients trading on international capital markets, and demonstrated superior performance in classifying illicit transactions compared to its LSTM and autoencoder counterparts. However, the evaluation focused on overall performance across both illicit and licit classes, limiting its effectiveness in detecting minority illicit transactions. Moreover, the dataset in~\cite{9758694} features a class imbalance ratio of 1:20, whereas real-world datasets often exhibit more severe imbalances, such as 1:5000.

In another study, \cite{xu2023illegal} introduces a novel approach to illicit account detection using a heterogeneous transformer network, referred to as AHGTN. This method constructs an account-centric network model on Ethereum, employs a graph transformer network to capture multi-hop paths and relationship metrics between account nodes, and incorporates a GCN for classification. While the model demonstrated superior performance compared to traditional machine learning techniques such as SVM and XGBoost, the dataset used comprised only 1,600 accounts with a 1:1 illicit-to-licit ratio, randomly sampled from the Etherscan and Etherscam databases. Hence, given the severe class imbalance and significantly higher transaction volumes typically encountered in practice, assessing the model’s effectiveness in real-world scenarios is challenging.

Furthermore, the authors\cite{lin2024fraudgt} build upon the graph neural network enhancement techniques proposed in \cite{egressy2024provablypowerfulgraphneural} to improve detection performance on the IT-AML dataset~\cite{altman2023realistic}. Specifically, the method involves constructing a subgraph centered on target transactions to form the sampled neighborhood. Techniques such as reverse message passing, port numbering, and Ego IDs are applied to enrich the subgraph features, which are then processed by the FraudGT encoder block to generate predictions. Experimental results show that leveraging two or more enhancement techniques enables FraudGT to outperform models like GAT, GIN, and PNA-based models. Nonetheless, the minority F1 score exhibits significant variation, ranging from 37\% for an illicit percentage of approximately 0.005\% to 77\% for an illicit percentage of around 0.11\%.

\begin{table*}[!t]
    \begin{center}
    \caption{A comparison of the GNN-based AML proposals}
    \begin{tabular}{|m{0.4cm}|m{1.35cm}|m{1.cm}|m{1.85cm}|m{2.5cm}|m{8.25cm}|}
        \hline
        \textbf{Ref.} &  \textbf{Dataset} & \textbf{Dataset size} & \textbf{Best model} & \textbf{Evaluation metrics} & \textbf{Insights} \\ \hline
        \cite{LaundroGraph} & Private & 33.3M & GNN & Average precision, ROC AUC  & By encoding the AML dataset into bipartite customer-transaction features and incorporating attributes such as risk rating and maximum transaction amount, GNN demonstrates superior performance compared to models like MLP, LightGBM, and DGI.\\ \hline
        \cite{pareja2020evolvegcn} & Elliptic  & 234,355 & GCN+GRU & Minority F1 & EvolveGCN computes a separate GCN model for each time step and connects them using an RNN. This shows that dynamic models are more effective in capturing and incorporating temporal information, thereby outperforming static GCN models.\\ \hline
        \cite{10.1145/3409073.3409080} & Elliptic & 234,355 & Modified GCN + linear layer  & Accuracy, precision, recall, F1 & The Modified GCN outperforms both the GCN and Skip-GCN models in terms of F1 score and overall accuracy. \\ \hline
        \cite{weberanti} & Elliptic  & 234,355 & EvolveGCN & Minority precision, minority recall, minority F1  &After comparing GCN, Skip-GCN, and EvolveGCN, it was concluded that EvolveGCN achieves the best minority F1 score, demonstrating the effectiveness of incorporating both graph structure and temporal information.\\ \hline
        \cite{lo2023inspection} & Elliptic & 234,355 & Inspection-L & Precision, recall, F1, ROC AUC & By integrating DGI with random forest models, Inspection-L outperforms traditional machine learning techniques, as well as GCN, Skip-GCN, and EvolveGCN.\\ \hline
        \cite{egressy2024provablypowerfulgraphneural} & IT-AML; ETH & 5-31M; 13M & GNN & Minority F1 & By leveraging enhancement techniques such as multigraph port numbering, ego IDs, and reverse message passing in GNN, the model achieves superior performance compared to GIN, PNA, and R-GCNs\\ \hline
        \cite{10.1007/978-3-031-33377-4_10} & Bitcoin dataset\cite{9332279}  & 14M-94M & Multi-relational GNN & Precision, recall, ROC AUC & By integrating edge orientations and characteristics into a multi-relational GNN model and combining it with an adaptive neighbor sampler, the model can identify similar illicit nodes. The results show that it outperforms existing methods such as GCN, GAT, and PC-GNN.\\ \hline
        \cite{10114503} & Private & 5.38M & GAGNN & Recall, ROC AUC & By encoding transaction information into graph embeddings and aggregating nodes with similar behaviors, organized laundering activities can be identified. Experimental results show that it outperforms GAT, GraphSAGE, Graphormer, GCN, GraphConsis, and PC-GNN. \\ \hline
    \end{tabular}
    \label{tab:GNNsum}
    \end{center}
\end{table*}

\subsection{GNN-based AML} \label{subsec: GNN}
GNN-based methods have demonstrated significant potential in detecting money laundering activities due to their ability to model complex relational and structural patterns in financial transaction data, which can be naturally represented as graphs~\cite{10477569}. Unlike traditional methods that analyze transactions or entities in isolation, GNNs leverage the relationships between nodes (e.g., accounts) and edges (e.g., transactions) to capture contextual dependencies, enabling the identification of complex laundering techniques such as circular transactions, smurfing, or layering. By learning node and edge embeddings that encode both local features and global graph properties, GNNs can uncover anomalies indicative of money laundering, such as abnormal transaction flows or high-centrality nodes. Additionally, GNNs can incorporate supporting data, such as customer profiles and temporal information, to enhance detection accuracy while offering explainability in certain architectures, which is critical for regulatory compliance.

One of the common applications of GNNs in AML is link prediction, where statistical metrics derived from money laundering activities are incorporated to improve detection~\cite{weberanti, LaundroGraph, oliveira2021guiltywalker}. For example, LaundroGraph~\cite{LaundroGraph} leverages customer profiles constructed from historical transaction data for transaction classification, incorporating features such as risk ratings and maximum transaction amounts. By encoding customers and transactions as graph representations using GNNs, the model transforms a confidential real-world banking dataset into a directed bipartite customer-transaction graph to predict anomalous links between customer and transaction nodes. Experimental results show that LaundroGraph outperforms other models, including MLP, LightGBM~\cite{lightgbm}, and DGI~\cite{veličković2018deepgraphinfomax}. However, the confidential and private nature of the dataset used for validation poses challenges in benchmarking its performance against other state-of-the-art deep learning models.

Another common application of GNN in AML is the classification task and numerous GNN-based approaches leverage the Elliptic dataset to develop methods for node or edge classification. These proposals aim to build effective predictive models capable of determining whether a specific account or transaction is licit or illicit~\cite{weberanti, pareja2020evolvegcn}.
For example, Pareja et al.~\cite{pareja2020evolvegcn} proposes EvolveGCN, an extension of GCN designed to capture the temporal dynamics inherent in blockchain transaction systems and perform node classification. This approach computes a separate GCN model for each time step and connects them using a RNN to account for system dynamics. While the minority F1 score of EvolveGCN and its variant outperforms static GCN models, it falls short compared to the GCN-GRU model, which co-trains a single GCN with a GRU. Furthermore, although experimental results indicate that dynamic models are more effective when incorporating temporal information from AML datasets, their minority F1 scores remain below 0.6, highlighting room for improvement in detecting minority illicit activities.

Likewise, \cite{10.1145/3409073.3409080} proposes an innovative method that integrates modified GCNs ~\cite{gasteiger2022predictpropagategraphneural} with linear layers to identify illicit nodes in the Elliptic dataset. Experimental results showed that this GCN-based approach achieved superior performance in terms of F1 score and overall accuracy compared to existing models such as GCN and Skip-GCN.

Similarly, Weber et al.~\cite{weberanti} evaluates the performance of the GCN, its variant Skip-GCN, and EvolveGCN~\cite{pareja2020evolvegcn} for binary classification of illicit transactions. To address class imbalance, they utilized a weighted cross-entropy loss function and adopted a 0.3-to-0.7 data ratio of licit to illicit transactions, ensuring the model placed greater emphasis on the minority class. Their results showed that GCN, Skip-GCN, and EvolveGCN outperformed logistic regression in terms of the minority F1 score, achieving 0.628, 0.705, and 0.720, respectively. These findings highlight the effectiveness of leveraging graph structures and incorporating temporal information in enhancing the classification of illicit transactions.

In addition, \cite{lo2023inspection} introduces a novel framework for detecting illicit transactions in the Elliptic dataset called Inspection-L. Inspection-L is a self-supervised, GNN-based approach that generates node embeddings from transaction graph patterns and node features without relying on labeled data. To capture both the topological structure and node features of transaction graphs, Inspection-L integrates the strengths of an enhanced self-supervised Deep Graph Infomax (DGI)\cite{veličković2018deepgraphinfomax} alongside a supervised Random Forest (RF)-based classifier. Experimental results show that Inspection-L significantly outperforms traditional machine learning methods, such as logistic regression, random forest, and XGBoost, as well as advanced deep learning techniques such as GCN, Skip-GCN, and EvolveGCN\cite{pareja2020evolvegcn}. Nevertheless, the framework’s highest recall score of 0.721 indicates that further improvements are needed to enhance its effectiveness in identifying illicit transactions.

Other variations of GNN-based approaches for AML classification tasks include the incorporation of novel technical adaptations~\cite{egressy2024provablypowerfulgraphneural} or the integration of group-awareness mechanisms to enhance the identification of illicit transactions~\cite{10114503}.

Egressy et al.~\cite{egressy2024provablypowerfulgraphneural} proposes a novel approach that combines technical adaptations with GNNs to detect money laundering patterns in transaction networks. Their method incorporates advanced techniques such as multigraph port numbering, ego IDs, and reverse message passing to identify complex subgraph patterns indicative of illicit activity. Using an AML dataset generated by a real-world simulator~\cite{altman2023realistic} and an Ethereum transaction network published on Kaggle, their model demonstrated superior performance compared to baseline models, including GIN~\cite{xu2018powerful,hu2020strategiespretraininggraphneural}, PNA~\cite{veličković2018deepgraphinfomax}, and R-GCNs~\cite{10.1007/978-3-319-93417-4_38}. However, the primary focus of their model was detecting subgraph patterns associated with money laundering, which presents limited feasibility for practical AML applications. For example, such tasks require constructing comprehensive transaction networks by aggregating data across banks and borders, a process constrained by data privacy concerns and banking regulations. Nevertheless, this approach offers valuable insights and a potential direction for cross-border financial crime investigations, particularly in scenarios where international collaboration facilitates the resolution of complex financial crimes.

Similarly, \cite{10.1007/978-3-031-33377-4_10} proposes a novel approach that integrates edge orientations and characteristics, such as transaction amounts and frequencies, into a multi-relational GNN model, combined with an adaptive neighbor sampler, to detect illicit transactions in a publicly available Bitcoin dataset\cite{9332279}. By adopting a multi-relational framework that incorporates both transactional and cooperative relationships as graph representations, the model is designed to identify nodes with similar attributes while effectively distinguishing illicit nodes. Experimental results reveal that the proposed model, BitcoNN, significantly outperforms existing methods, including GCN, GAT~\cite{gat2018graph}, and PC-GNN~\cite{10.1145/3442381.3449989}, demonstrating its effectiveness in detecting illicit activities within cryptocurrency networks.

While in \cite{10114503}, the authors investigate group-level interactions in money laundering schemes, where criminals exploit groups of accounts to layer funds and obscure illicit activities. They propose a novel model called Group-aware GNN (GAGNN), which encodes transaction information into graph embeddings and incorporates a scheme for aggregating nodes with similar behaviors to learn patterns of organized activities. The model also utilizes down-sampling and optimization strategies specifically designed for classification tasks. Experiments conducted on a large-scale real-world dataset collected from UnionPay reveal that GAGNN significantly outperforms existing models, including GAT~\cite{gat2018graph}, GraphSAGE~\cite{hamilton2018inductiverepresentationlearninglarge}, GCN, Graphormer~\cite{NEURIPS2021_f1c15925}, GraphConsis~\cite{10.1109/TSE.2024.3385538}, and PC-GNN~\cite{10.1145/3442381.3449989}, in detecting both individual and organized suspicious money laundering transactions. However, in addition to the common limitation of private datasets, the authors acknowledge that the results may be influenced by the current data distribution and patterns, potentially limiting the model's robustness in adapting to dynamically evolving money laundering strategies. To summarize, Table \ref{tab:GNNsum} presents a comparison of GNN-based models along with key insights derived from these proposals.

\subsection{DRL-based AML}

\begin{figure}[hb!t]
\begin{center}
    \includegraphics[width=0.35\textwidth]{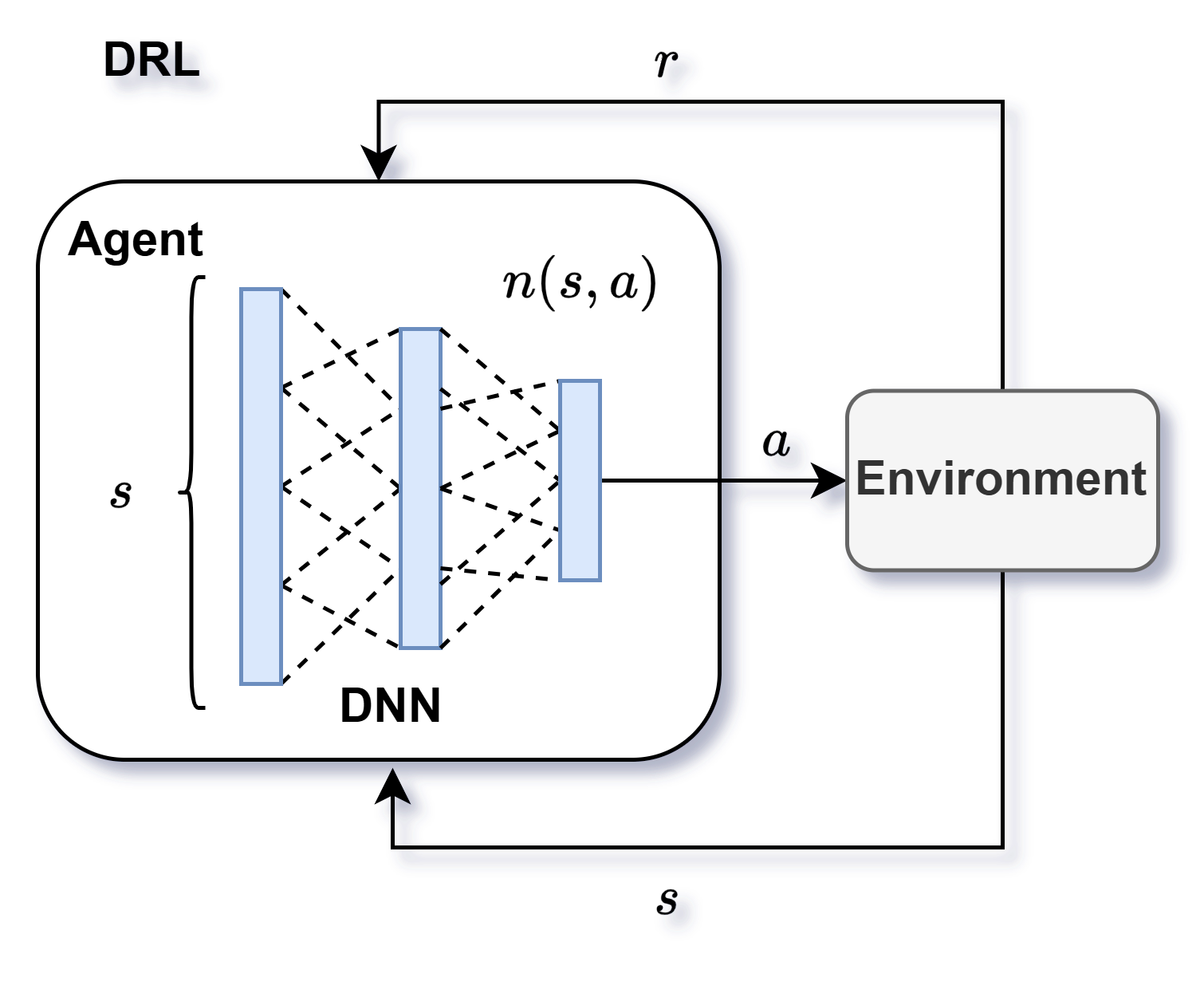}
    \caption{\textcolor{black}{An illustration of a typical DRL system comprises an intelligent agent trained using a DNN to perform actions \( a \), observe states \( s \), and receive rewards \( r \) using policy \( n(s,a) \).}}
    \label{fig:DRL}
\end{center}
\vspace{-5mm}
\end{figure}

DRL is highly suitable for money laundering detection because it combines the strengths of RL with the powerful representation capabilities of deep learning, making it effective in handling the complexity and dynamism of money laundering activities. \textcolor{black}{As shown in Fig. \ref{fig:DRL},} DRL can process large-scale, high-dimensional transaction datasets by leveraging deep neural networks to extract complex features and patterns, which traditional RL methods might struggle to capture~\cite{10272628}. Money laundering often involves sequential decision-making processes, such as identifying unusual patterns across multiple stages of transactions, i.e., placement, layering, and integration. 

Implementing DRL often requires defining an optimal policy and explicitly specifying the observations for the learning agent~\cite{8102446}, which demands substantial domain expertise and is less automated compared to other deep learning solutions for AML. Nonetheless, a few DRL-based proposals, such as ~\cite{vimal2021applicationdeepreinforcementlearning}, have been explored for financial fraud detection.

In \cite{vimal2021applicationdeepreinforcementlearning}, the authors formulates the fraud detection problem as a DRL task and introduced a novel tunable reward function designed to maximize revenue while maintaining a balance between the fraud rate (i.e., the percentage of fraudulent transactions approved by the agent) and the decline rate (i.e., the percentage of legitimate transactions declined). Using the Markov decision process (MDP) framework, the authors defined the state as the feature vector of a transaction, with two possible actions available to the deep q-network (DQN)\cite{mnih2015human} agent, i.e., approve or decline the transaction. The agent is rewarded for correctly rejecting fraudulent transactions and penalized for incorrectly approving them. By testing various reward functions on the European card data (ECD)\cite{DalPozzolo2015} and the IEEE-CIS fraud dataset\cite{kaggle_ieee_fraud_detection}, the proposed DQNR model is able to achieve performance that is comparable to or better than other models, including CNN, LSTM, and XGBoost. Although the paper primarily focuses on credit card fraud detection rather than AML predictions, it presents a potential direction for leveraging DRL techniques in AML datasets.

In conclusion, various deep learning approaches offer distinct advantages over traditional machine learning techniques in AML tasks, demonstrating superior performance in handling complex patterns and large-scale financial datasets. However, these models are not without limitations, as several challenges persist that may affect their applicability and robustness in real-world scenarios. In the next section, we will examine the challenges commonly encountered in deep learning-based AML solutions and explore potential strategies to address them.

\section{Challenges for Deep Learning Application in AML} \label{sec: challenges}

\begin{figure}[!t]
\begin{center}
    \includegraphics[width=0.35\textwidth]{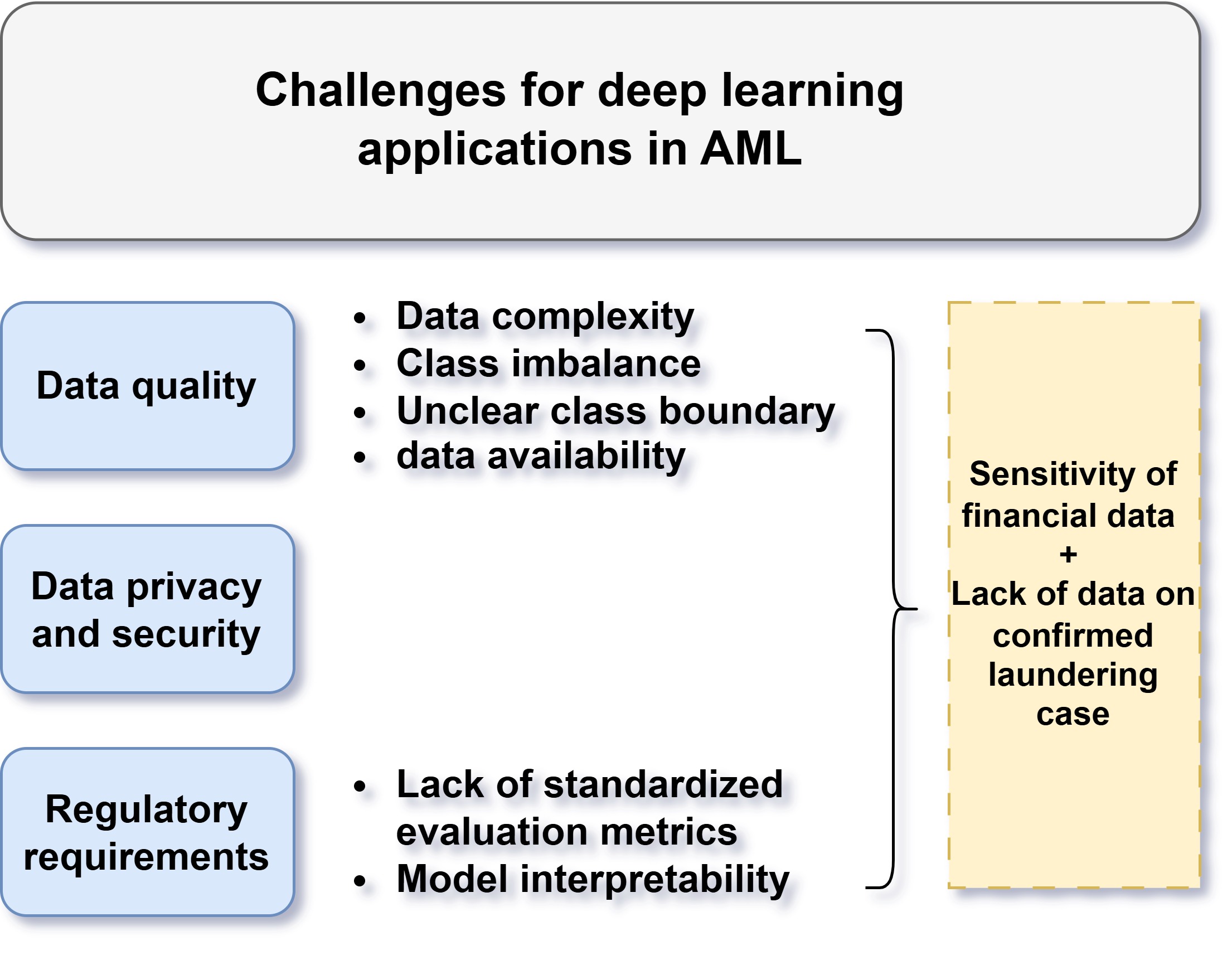}
    \caption{A summary of the common domain-specific challenges encountered by deep learning approaches in the application of AML.}
    \label{fig:challenge}
\end{center}
\vspace{-5mm}
\end{figure}

\textcolor{black}{In the context of deep learning-based AML, systems often need to process a heterogeneous pool of datasets that capture various aspects of user financial behavior and money flow. While the availability of multivariate data can enhance the understanding of transaction contexts and improve decision-making accuracy, it also increases the complexity of data processing and places greater demands on the model’s ability to extract and leverage relevant features effectively~\cite{10.1007/s42979-021-00592-x}. Additionally, many regulatory decisions regarding the validity of money laundering suspicion reports are not shared with financial institutions, resulting in a scarcity of positive money laundering records in datasets. This leads to significant class imbalance during model training~\cite{10.1007/s10115-017-1144-z}. The issue of data quality is further exacerbated by class overlaps and concept drift. These factors create additional challenges, making detection even harder in the absence of clear classification boundaries. Therefore, this section will explore in depth the challenges faced by deep learning applications in AML to highlight domain-related issues and discuss potential solutions to address them. A summary of these challenges are presented in Fig. \ref{fig:challenge}.}


\subsection{Data Quality}

Data quality in machine learning is broadly assessed using key metrics, such as accuracy, correctness, completeness, and relevance~\cite{jordan2015machine}. Specifically, data accuracy ensures that the sampled data reflects actual trends, with validation measures taken to verify its reliability. Data correctness focuses on the absence of errors, inconsistencies, or noise during data collection, transfer, or storage, such as unprocessed duplicate entries. Data completeness evaluates whether the data provides sufficient information to capture the overall statistics of the observed events or individuals. Meanwhile, data relevance examines whether the data is relevant to the current task; excessive irrelevant or redundant data can lead models to memorize the training data rather than learning underlying patterns, thereby affecting their ability to generalize to unseen data~\cite{wang1996beyond}.

Data quality influences model performance, particularly in tasks requiring high detection accuracy. Models trained on data containing errors, inconsistencies, or excessive noise often learn incorrect patterns, making it harder to extract meaningful features and resulting in lower predictive performance. Consequently, detection accuracy is closely coupled with data quality assurance, which systematically investigates and addresses desirable data attibutes. Without such assurance, achieving high detection accuracy is nearly impossible.

For deep learning applications in AML, some critical aspects of data quality that frequently undermine model performance are:

\subsubsection{Data Complexity}
Data complexity poses a significant challenge in training deep learning models for AML applications, stemming from various factors that complicate data processing and management. These include the massive volume of transaction records generated digitally across interconnected payment platforms, the high velocity of data generation facilitated by mobile payments, and the diverse formats and sources of collected data.

A major contributor to data complexity in AML systems is the sophisticated and evolving techniques used by criminals to evade detection~\cite{9964052}. AML systems must not only learn from historical laundering patterns but also accurately identify modified or emerging fraud patterns. This poses a serious challenge to the robustness of detection models and their ability to adapt to new laundering techniques. Furthermore, validating suspicious transaction alerts generated by rule-based systems often requires integrating diverse data from multiple sources and historical records. For example, AML teams rely on information gathered through KYC procedures, past transaction histories, analyses of financial relationships, and intelligence from regulatory bodies. This multifaceted and heterogeneous data is essential for verifying the legitimacy of flagged transactions, highlighting the need for advanced AML systems capable of handling such data complexity effectively.

Incorporating diverse data sources into deep learning models can enhance performance by providing features and a rich context to describe each transaction. However, the complexity of processing this data and integrating it into a cohesive and clean input set for training presents considerable challenges. These challenges include handling heterogeneous data formats, ensuring consistency across sources, and addressing potential data quality issues~\cite{OZTAS2024161}. Furthermore, the sharing of sensitive financial information across parties raises significant privacy concerns, which limit data availability and further constrain the potential accuracy that the model could achieve. Overcoming these barriers requires sophisticated data preprocessing pipelines and robust privacy-preserving techniques to balance model effectiveness with data security.

\subsubsection{Class Imbalance}
Class imbalance occurs when certain classes in a dataset are significantly underrepresented, often due to the rarity of minority class events in real-world scenarios or inadequate sampling methods~\cite{johnson2019survey}. In datasets with highly skewed ratio of majority to minority class records, models tend to become biased toward the majority class during training. This bias causes models to focus on patterns from the majority class while neglecting the minority class, which is often the most critical for the task. Consequently, models fail to accurately classify the minority class and struggle to generalize effectively to unseen data, particularly when the minority class represents rare but important events.

This issue is particularly critical in fraud detection tasks, such as AML applications and anomaly detection~\cite{10486886}. These domains generate millions of transaction records per second, with the vast majority being legitimate. Fraudulent or suspicious transactions—the minority class—are rare but essential to detect. Without addressing class imbalance, models may achieve high overall accuracy but perform poorly in identifying fraudulent activities, undermining the primary goal of AML systems. Moreover, the imbalanced nature of the data exacerbates problems related to high false positives and false negatives. False positives, where legitimate transactions are flagged as suspicious, can inflate operational costs, reduce system efficiency, and disrupt customer experiences during AML investigations. On the other hand, false negatives pose severe risks to financial institutions, including financial losses, reputational damage, and regulatory penalties.

In addition, class imbalance in AML datasets poses significant challenges for evaluating research proposals in this domain. Validating the performance of proposed methods typically requires access to real datasets or widely recognized public datasets to assess their applicability in real-world scenarios~\cite{vogelsgesang2018get}. However, obtaining data on confirmed cases of money laundering from regulatory agencies is highly challenging due to local restrictions and stringent data privacy regulations. As a result, institutional datasets often suffer from severe class imbalance, consisting primarily of suspected cases rather than confirmed instances of money laundering or undetected fraudulent activities that escaped detection.

\textcolor{black}{To address this, many private datasets used for model training and validation incorporate artificially generated positive cases to simulate money laundering instances~\cite{10.1007/s10115-017-1144-z}.} While this approach partially mitigates the lack of minority class data, it introduces inherent limitations. Simulated data may fail to capture the complexity and variability of authentic money laundering patterns, undermining the reliability of model evaluation and reducing the generalizability of proposed methods to real-world AML systems. This discrepancy highlights the reliability of reported results and hampers the generalizability of proposed methods in effectively detecting actual money laundering activities, thereby diminishing their feasibility in real-world AML systems.

Addressing class imbalance requires specialized techniques to ensure that the minority class is adequately represented and that models can effectively learn its patterns. One commonly used approach is resampling, which involves either oversampling the minority class to create more balanced data or undersampling the majority class to reduce its dominance. Cost-sensitive learning is another effective strategy, where the loss function can be adjusted to assign higher penalties to misclassifications of the minority class, compelling the model to focus on accurately classifying these critical instances~\cite{10496993}. Additionally, advanced algorithmic techniques, such as generative models, can be employed to synthesize realistic samples of the minority class, further mitigating the imbalance while preserving data authenticity.

To evaluate the performance of AML models under realistic conditions, open AML datasets such as the Elliptic dataset~\cite{elliptic_dataset} and the IT-AML dataset~\cite{altman2023realistic} play a crucial role. These datasets simulate real-world financial transactions and money laundering patterns, offering a representative and practical framework for testing and validation. By leveraging these resources, researchers can effectively assess the robustness and generalizability of their models. Such resources ensure that models are not only capable of addressing class imbalance comprehensively but are also highly relevant to practical AML applications.

\subsubsection{Unclear Class Boundary}
Unclear class boundaries pose a significant challenge to maintaining data quality in AML systems~\cite{smith2014instance}. Due to strict banking laws and data protection regulations, sensitive financial information is heavily safeguarded and often inaccessible to the public or even financial institutions. Although banks routinely file suspicion reports to flag potential money laundering cases, they frequently lack access to the final legal outcomes that confirm or refute these suspicions. This information gap creates significant uncertainty classifying financial activities~\cite{10.1007/s10115-017-1144-z}.

Even when crimes are publicly disclosed, identifying the specific transactions within a suspect's account that are directly tied to money laundering remains an arduous task. Suspects may have conducted thousands of transactions over years, with only a small subset potentially linked to illicit activity. Without clear labels or definitive legal records, it becomes exceedingly difficult to distinguish legitimate transactions from those associated with criminal behavior. 

This lack of clarity not only undermines the accuracy of AML systems but also strains investigative resources and reduces the efficacy of detection efforts. Addressing this issue will require enhanced collaboration between governing bodies, financial institutions, and the research community. Such initiatives can improve the accuracy and reliability of AML systems, strengthening their ability to detect and mitigate financial crimes effectively.

\subsubsection{Data availability}
A significant challenge in applying deep learning models to AML is the scarcity of real-world datasets with clearly labeled money laundering cases. Regulatory frameworks, including banking and data privacy laws, impose strict restrictions on the collection, sharing, and use of customer financial data to protect sensitive information~\cite{10.1007/s10115-017-1144-z}. While these safeguards are essential for maintaining client trust and ensuring compliance, they inadvertently create obstacles to the development and evaluation of effective AML systems.

Deep learning models rely on large, diverse, and accurately labeled datasets to detect complex patterns and anomalies~\cite{9763022}. However, in the AML domain, the highly confidential nature of financial data and the associated legal restrictions severely limit data availability~\cite{amldisclosure}. Without access to real-life datasets, researchers often resort to using synthetic or simulated data, which may not accurately reflect the evolving nature of real-world money laundering activities.

Furthermore, many proposed AML models are trained and tested on private datasets that remain inaccessible to the broader research community~\cite{9446887}. This lack of transparency not only hinders reproducibility but also renders the results less useful for comparison. Without a common benchmark dataset, researchers face a fragmented landscape where validating the efficacy of various AML solutions becomes increasingly challenging.

To address the challenge of AML data availability, greater collaboration between regulatory bodies, financial institutions, and the research community is essential. Initiatives such as secure data-sharing frameworks, federated learning systems, and anonymized datasets could help bridge this gap. Additionally, open AML datasets that simulate real-world financial transactions and money laundering patterns, such as the Elliptic dataset~\cite{elliptic_dataset} and the IT-AML dataset~\cite{altman2023realistic}, can serve as valuable validation tools to establish the adaptability and robustness of AML models.

\subsection{Data Privacy and Security}

Existing AML systems often face significant inefficiencies when processing the large volume of flagged suspicious activities within traditional database structures~\cite{LaundroGraph}. This inefficiency results in prolonged investigation periods, during which suspicious activities may remain undetected. Consequently, the success rate in identifying and preventing money laundering is often low, undermining the effectiveness of AML efforts.

Beyond technical inefficiencie, AML investigations involve the analysis of highly sensitive client financial data, making information security a critical concern. Ensuring the confidentiality, integrity, and availability of this data is essential to safeguard stakeholder interests and maintain trust in the financial system~\cite{thommandru2023exploring}. To achieve this, it is crucial for AML teams to implement strict access controls, ensuring that only authorized personnel can access the data necessary for investigations. Unauthorized access may lead to severe privacy breaches, identity theft, financial espionage, and expose clients to risks such as phishing and social engineering attacks. For example, criminals may exploit individuals' donation patterns by posing as legitimate charitable organizations or creating fraudulent campaigns, deceiving users accustomed to making contributions. Such deception preys on clients' trust and generosity, making it difficult to distinguish legitimate activities from fraudulent ones. 

Moreover, safeguarding the integrity of financial data is equally crucial. Any compromise in data integrity can lead to incorrect conclusions, impeding the detection of money laundering and potentially resulting in wrongful actions against innocent clients. For example, errors in transaction data may trigger false positives, causing legitimate activities to be flagged as suspicious. This not only triggers unnecessary investigations but also causes significant inconvenience to clients. Equally critical is ensuring the availability of sensitive data during investigations~\cite{10058199}. Delays or disruptions in accessing this information can hinder the timely identification and prevention of illicit activities, leaving money laundering undetected. As client data is shared across various departments, third-party vendors, and regulatory bodies, the increasing volume and variety of sensitive information highlight the pressing need for stringent data protection measures to ensure both security and compliance.

In addition, protecting the privacy of sensitive financial data is a legal and regulatory imperative. Beyond legal compliance, data privacy breaches can expose the institution to significant legal, regulatory, and reputational risks~\cite{10058199}. Non-compliance with data protection laws, such as GDPR or the California Consumer Privacy Act (CCPA)\cite{oagccpa}, could result in severe penalties. Furthermore, the erosion of client trust due to inadequate data protection may irreparably damage long-term relationships. As clients become increasingly aware of data privacy concerns, institutions that fail to adequately protect sensitive information risk losing customers and facing lasting financial repercussions.

To mitigate these risks, AML systems must incorporate stringent data protection measures. This includes implementing strong access controls to ensure that only authorized personnel can access specific types of sensitive data and utilizing advanced encryption technologies to secure data both in transit and at rest. Additionally, adopting techniques such as differential privacy or federated learning~\cite{9830997}, which allow data analysis without direct access to sensitive data, can enhance security and minimize the potential for privacy breaches.

The ultimate objective of any AML system is to strike a balance between effectively detecting and preventing fraudulent activities with the privacy and security of innocent clients. A robust, secure, and efficient AML system not only helps financial institutions protect themselves from financial crime but also ensures that clients' data is treated with the utmost care and respect. By addressing these challenges, AML systems can maintain their integrity, build trust with clients, and contribute to the overall security of the financial system.

\subsection{Regulatory Requirement}

Money laundering often originates from predicate crimes, such as fraud, drug trafficking, and corruption, where the illicit proceeds stem from offenses\cite{amlandterrorism}. These predicate offenses form the foundation of the money-laundering process, making financial crimes a top priority for governments. In response, stringent laws and comprehensive preventive measures have been enacted to detect, disrupt, and deter these illicit activities effectively.

AML regulations focus on tracing the origins of illicit funds and monitoring suspicious transactions that may signal money laundering attempts~\cite{thommandru2023exploring}. By targeting the laundering process at its inception and throughout its lifecycle, these measures aim to dismantle the financial networks supporting criminal enterprises. Financial institutions, particularly major banks, play a central role in upholding AML standards by conducting rigorous due diligence, employing sophisticated monitoring systems, and ensuring their operations remain compliant with regulatory frameworks.

Failure to meet AML obligations carries severe repercussions. Non-compliant institutions face substantial fines, legal penalties, and reputational damage. Furthermore, investigations triggered by non-compliance can lead to the freezing of client assets, eroding trust and undermining client confidence. Such breaches not only harm individual clients but also pose a broader threat to the integrity of the financial system. To mitigate these risks, AML systems must be carefully designed to adhere to regulatory requirements while ensuring both compliance and operational effectiveness. Key regulatory requirements for AML systems include~\cite{fatf_virtual_assets}:
\begin{itemize}
    \item Verification and collection of customer information: KYC procedures~\cite{gill2004preventing} are conducted during account opening and through ongoing monitoring of customer activities. These processes involve handling heterogeneous and highly confidential data, requiring extreme caution in its processing and analysis to ensure compliance and maintain trust.
    \item Transaction monitoring: Financial institutions must continually update their detection models to adapt to evolving money laundering techniques and societal behavior norms, ensuring the sustained quality of detection~\cite{10356200}. Additionally, to evaluate the performance of AML systems effectively, a standardized evaluation metric is essential for tracking and validating their outcomes.  
    \item Suspicious activity reporting: AML systems are required to promptly document and report suspicious transactions to regulatory authorities. This requires high model interpretability to provide sufficient evidence supporting flagged suspicions, ensuring that the reporting process is both reliable and actionable.  
    \item Data privacy and security: Protecting sensitive financial data from misuse, unauthorized disclosure, and breaches is important. AML systems must implement robust privacy and security measures to comply with data protection regulations and maintain the integrity of client information.  
\end{itemize}

Besides constraints on data availability, two other significant factors hindering the adoption of deep learning in the AML domain are the lack of standardized evaluation metric and the interpretability of deep learning models.

\subsubsection{Lack of standardized evaluation metric}
The absence of standardized evaluation metrics is a significant obstacle to advancing deep learning applications in AML. Most AML datasets exhibit significant class imbalances, with legitimate transactions greatly outnumbering illicit ones. Hence, overall accuracy becomes an unreliable performance measure, as models tend to favor the majority class, producing inflated accuracy scores that fail to reflect their true effectiveness in detecting illicit activities~\cite{10025710}.

To address this challenge, adopting alternative evaluation metrics is crucial. Metrics such as the receiver operating characteristic (ROC) curve,  and the area under the ROC curve (AUC-ROC) provide a comprehensive evaluation of a model's discriminatory ability~\cite{zhou2021evaluating}. Additionally, precision, recall, and the F1-score, particularly for the minority class, are essential for assessing a model's capability to accurately identify suspicious transactions while minimizing false positives and false negatives~\cite{app12199637}. 

\begin{figure*}[hb!t]
\begin{center}
    \includegraphics[width=0.65\textwidth]{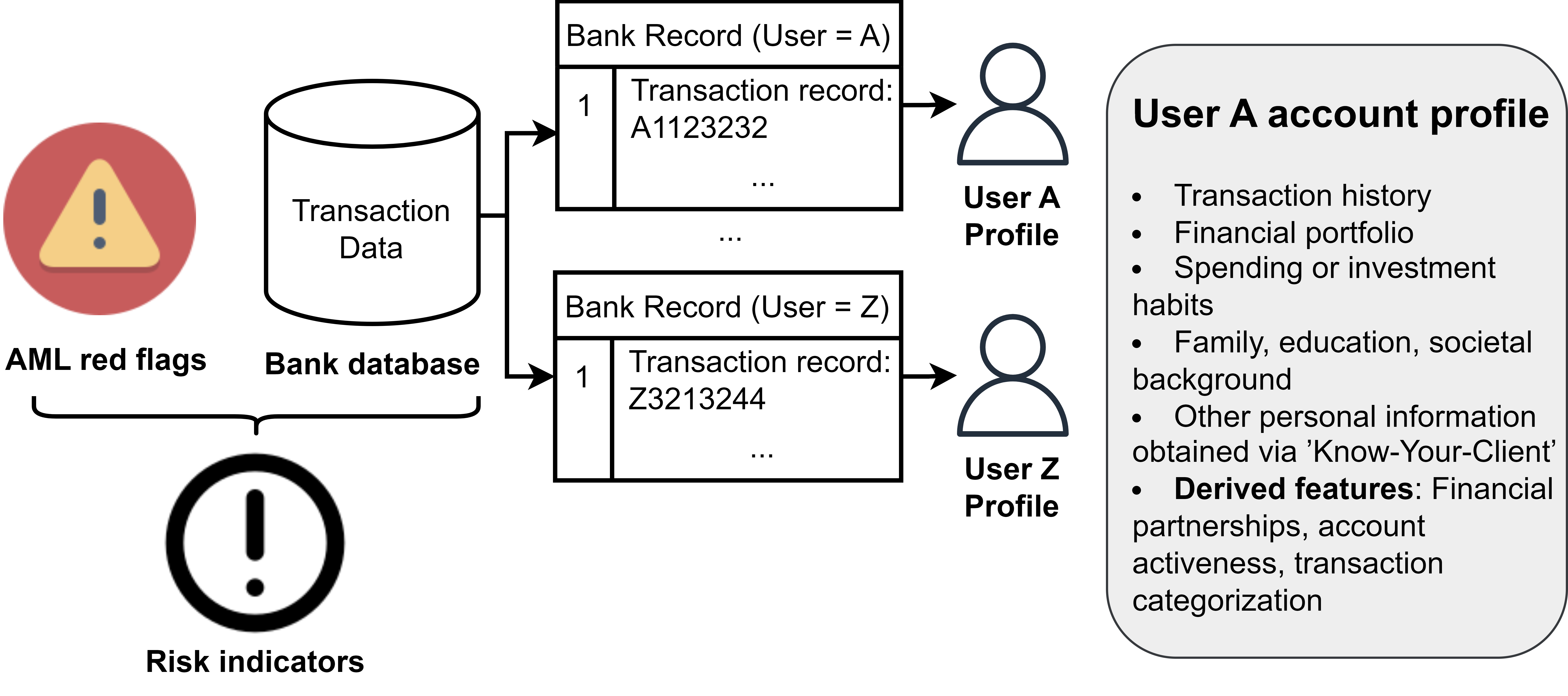}
    \caption{An illustration of the data preprocessing steps, which include conducting account profiling and generating risk indicators based on domain knowledge and statistical insights to contextualize each transaction. These are then combined with the corresponding transaction record as inputs for the prediction model.}
    \label{fig:account}
\end{center}
\vspace{-8mm}
\end{figure*}

\subsubsection{Model interpretability}
Regulatory frameworks, such as GDPR\cite{GDPR_2022} and CCPA \cite{oagccpa}, impose strict requirements for the interpretability and explainability of deep learning models, particularly in regulatory compliance contexts. In AML efforts, when institutions file suspicion reports (SARs), they must provide concrete and verifiable evidence to justify the suspicion. However, the inherent black-box complicates transparency, making it challenging to explain predictions in a clear and understandable manner~\cite{10756563}. This lack of interpretability presents a significant barrier, as regulatory bodies require institutions to justify their decisions with clear, understandable reasoning. Without sufficient transparency, deep learning models may fail to meet the compliance standards necessary to ensure legal and operational integrity in financial institutions~\cite{10021108}.

To build trust and enhance the efficiency of deep learning in AML, explainable artificial intelligence (XAI)~\cite{10509682} offer a promising solution. XAI provides visibility into how models arrive at prediction, enabling institutions to obtain the legal support required for the suspicion reports~\cite{gilpin2018explaining}. For example, a review by~\cite{9446887} highlights various deep learning techniques and potential XAI approaches applied to money laundering detection, showcasing how these technologies can be used to meet regulatory expectations while enhancing model transparency.

To address some of the aforementioned challenges, including privacy concerns surrounding sensitive financial information and the constraints of limited data availability, we propose a novel framework that integrates the principle of least privilege with deep learning techniques. This framework is tailored to support AML investigators in detecting fraudulent activities under restricted data access conditions.

\section{Overview of Our Approach} \label{sec: overview}

The main objective of our approach is to enable the efficient and effective identification of money laundering activities under real-world constraints. Banks typically have limited access to cross-bank financial data, as they can only access their local datasets and cannot track the flow of transactions into other banks. This limitation makes predicting money laundering patterns challenging. Additionally, safeguarding the privacy of sensitive financial information is a crucial requirement under banking laws. Therefore, minimizing privacy exposure for innocent users during AML activities is equally important.

In practice, machine learning and codified red flags are the two most commonly used methods for AML. While machine learning techniques can provide individualized predictions for each transaction record, the quality of these predictions often relies heavily on the input data, potentially missing the broader context and resulting in a high false alarm rate~\cite{10671571}. Conversely, a detection system based solely on broad domain knowledge focuses on established theories and logic, which tend to be inflexible towards emerging fraud patterns and prone to raising excessive false positives due to overgeneralization. Therefore, in this paper, we aim to leverage the strengths of both the context-rich predictions from machine learning models and the codified AML domain expertise to achieve more accurate detection results with a lower false positive rate.

To address these challenges, we propose the CRP-AML framework, which implements the principle of least privilege for effective fraud detection under restricted data availability. With reference to Fig.~\ref{fig:account}, financial account profiles are built for every user account using past transaction records, which serve as the context for evaluating transactions. This profile can also be complemented with information gathered from KYC procedures~\cite{gill2004preventing}, such as educational background, family information, and other personal details. AML red flags are combined with statistical insights from bank databases and codified into risk features of each transaction, which highlight AML-relevant characteristics in transactions. To further reduce false alarms, the overall risk indicator values are also used to verify the validity of the predictions from our model.

During standard operations, each transaction is evaluated by a machine learning model that examines both the sender's and receiver's profiles, along with associated risk features, to assess the probability of money laundering activity. In the context of AML investigations, only pertinent account profiles and transactions involving the suspected account are made available to the AML team, either through direct interactions or substantial financial relationships. This restricted access minimizes privacy risks for uninvolved clients, as the AML team is not granted direct access to the complete transaction database.

In this section, we will introduce the dataset utilized in our analysis, and a detailed discussion of our CRP-AML model is presented in the next section.









\subsection{Dataset}
The dataset utilized for experimentation in this paper is the IT-AML dataset~\cite{altman2023realistic}. This synthetically generated dataset models a complex virtual world comprising entities with complex and intricate relationships. Each transaction record in the dataset includes 11 columns of data, such as timestamp information, details of the outgoing and incoming banks, associated bank accounts, amounts received and sent in the respective currency, and the payment format (e.g., cheque and credit card).

We chose this dataset because it closely resembles real-world statistics, capturing transactions throughout the entire money laundering cycle. For instance, the IT-AML dataset models all three phases of money laundering:
\begin{enumerate}
 \item Placement of funds: This involves obtaining funds through various sources of criminal activity, such as loan sharking, gambling, drug-related activities, etc. These activities vary in terms of the amount and frequency of money received.
 \item Layering of illicit funds: This phase involves integrating illicit funds into the financial system through activities such as deposits and investments.
 \item Integration of funds: This involves spending illicit funds by paying employees, purchasing supplies, transferring funds to nations with loose regulatory checks, and more.
\end{enumerate}
\textcolor{black}{Furthermore, various money laundering patterns have been constructed to demonstrate the complexity of money laundering networks. These patterns include fan-in/fan-out, gather-scatter, cycles, multi-partite, random, and stack patterns.}

\section{Method} \label{sec: method}

\begin{figure*}[hb!t]
\begin{center}
    \includegraphics[width=0.8\textwidth]{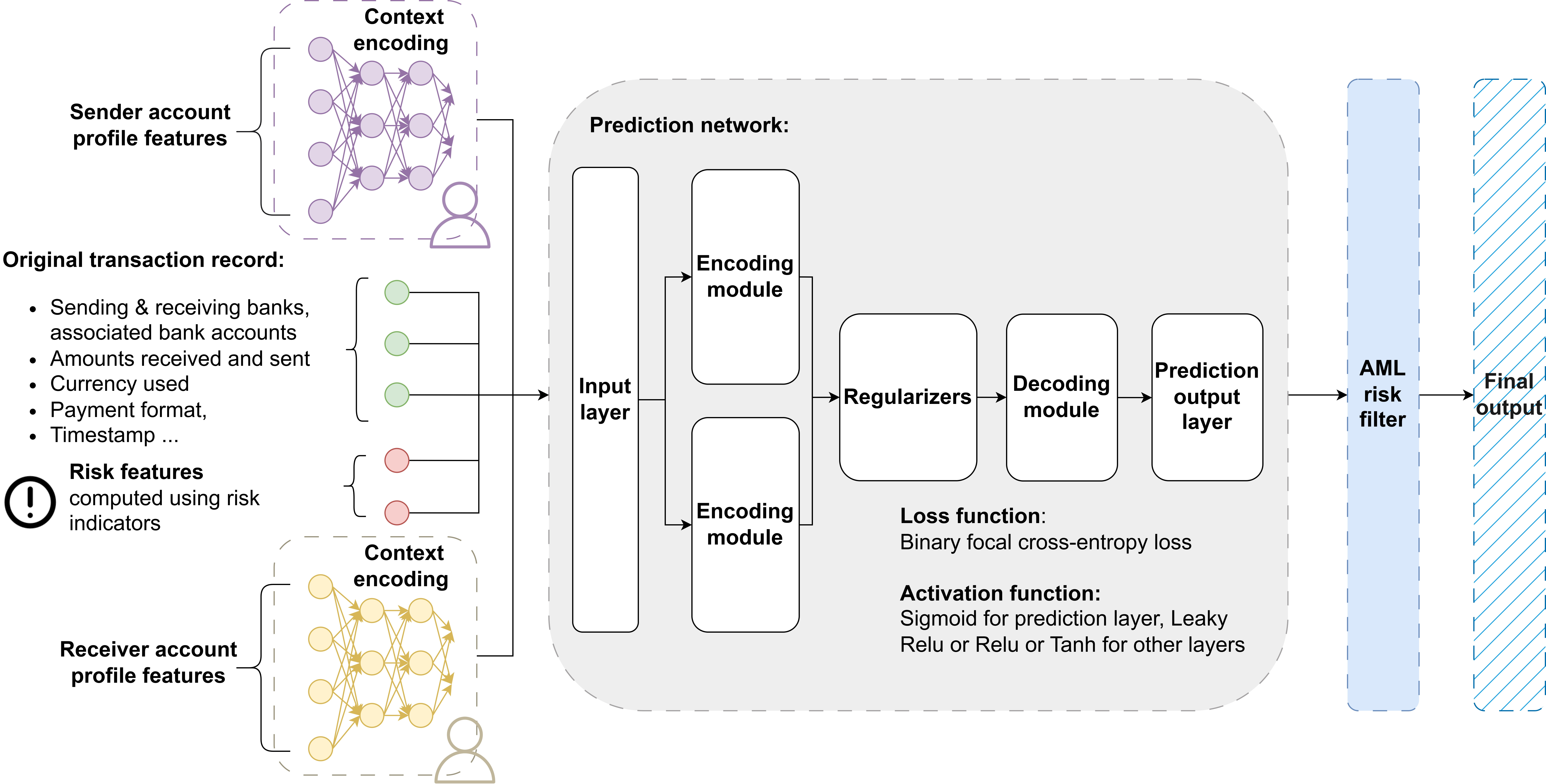}
    \caption{\textcolor{black}{An illustration of the general workflow of the CRP-AML model is presented. Contextual features (account profiles) are integrated with domain knowledge (risk features) and transaction records, which collectively serve as inputs to the model. After the prediction is generated, domain knowledge is again employed to filter out false positives, resulting in the final output.}}
    \label{fig:model}
\end{center}
\vspace{-8mm}
\end{figure*}

In this section, we provide a detailed explanation of how financial profiles are constructed for the accounts involved in the transaction database and how these profiles are incorporated as contextual embeddings to predict money laundering activities. We also explain how the knowledge of the AML domain is codified in various AML risk indicators that guide the accurate prediction of money laundering transactions. Following this, we introduce our proposed Context-Risk-Predict AML (CRP-AML) model for the automatic prediction of money laundering transactions and discuss the experimental results.

\subsection{Financial Account Profiling}
\textcolor{black}{Conventional financial fraud detection policies, such as the standard AML red flags used by banks, have limited ability to adapt to the specific behavioral characteristics of individual accounts, resulting in high false alarm rates~\cite{EDGE2009381}. To accommodate AML policies to individual account properties, comprehensive financial account profiles can be built for local clients that characterize user activities through behavior features, verify false alarms, and identify suspicious transactions that deviate from the behavior norm~\cite{7277068}. For example, the sender and receiver's financial account profiles can serve as the context for the transaction, and the risk of fraud associated with it is determined by the likelihood of its happening within this context.}

\textcolor{black}{Financial account profiling is a specialized form of user behavior analysis designed to understand accounts' financial patterns or signatures. It establishes a statistical representation of the financial behavioral norm. A typical financial account profile includes transaction history, financial portfolio, spending or investment habits, financial relationships, and other financial preferences associated with the account. These profiles are typically established KYC procedures, where banks verify customers and their financial backgrounds, as well as through the account's historical financial activities. In our experiments, we derive the financial account profiles statistically from historical transaction data, as no additional background information about clients is provided in the dataset. Specificcally, we first calculate the following account characteristics by data analysis techniques:
\begin{itemize}
    \item \textbf{Financial partnerships:} The context of an account can be understood as the result of its interactions with other accounts and its underlying financial objectives~\cite{cui2021remember}. For example, an account conducting periodic large-sum payments to another business account is likely engaged in a business partnership, while numerous small payments from many different customers combined with periodic large-sum payouts could characterize retail store operations. Therefore, we can describe the nature of the account in terms of its financial interconnections with others: the total number of unique transaction partners, the total transaction amount, the average transaction amount and frequency per partner, and the average time between each subsequent transaction with the same partner.
    \item \textbf{Account activeness:} The overall transaction statistics can effectively measure the account's activity level~\cite{7277068}. A large volume and high frequency of transactions indicate that the account is actively used by the account holder and can provide significant insight into the holder's overall financial behavior, thus yielding more conclusive results about the nature of the account. This includes features such as the total and average transaction volumes and the average transaction frequency of the account.
    \item \textbf{Transaction categorization:} Transaction methods, such as the major modes of payment and currencies involved, can convey important contextual information about the primary business function and location of the account~\cite{8735623}. For example, an account that performs only USD-based transactions using credit cards and cash is likely situated in locations where USD is the domestic currency and is likely a client-facing retail business serving individual customers rather than an investment firm. By coupling this information with a categorization of transaction amounts, we can infer the major types of transactions performed by the account, thereby establishing its behavior norm. 
    In our dataset, we classify each transaction by:
    \begin{itemize}
        \item The paying and receiving currency of the transaction: USD, Bitcoin, Australian Dollar, Yuan, Rupee, etc.
        \item The mode of payment of the transaction: Cheque, credit card, cash, and more.
        \item The amount (in terms of USD) involved in the transaction: amounts less than or equal to 50\% of all transactions (e.g., 1,000 USD) are considered small, 50-80\% (e.g., 10,000 USD) as medium, 80-93\% (e.g., 100,000 USD) as large, and anything beyond that as extra large.
    \end{itemize}
    For example, a retail store account might have 80\% ``small amount USD-based transactions via cash or credit card" and 20\% ``medium amount USD-based transactions via cash or credit card." In real banking systems, such transaction categorization can be further expanded to include the likely purpose of the transaction by utilizing registration details from business accounts. For example, a payment from an individual account to a utility company's business account is likely for a utility bill payment, among other possibilities.
\end{itemize}
}

For detecting money laundering events, an important indicator is the inconsistency of transactions with the financial profile of the account holder\cite{fatf_virtual_assets}. There are two types of account profiling feature we use to establish an account's norms, i.e., (1) past transaction records, which reflect an individual's daily spending patterns and identify common sources of funding; (2) the nature or classification of the account, such as whether it belongs to a high-net-worth individual, a large retail business, or a foreign exchange company. Accounts within the same classification exhibit similar behaviors or features. For example, a business account with an average net flow of 1 million typically represents a large company, where expected behaviors include paying large sums in taxes and handling large transaction amounts or volumes.

Thus, an extremely high-value retail purchase made using an account with a low-income categorization raises suspicions of financial scams or fraud. This suspicion is further compounded by sudden large deposits and a transaction history of non-luxury spending behavior. In contrast, such a purchase would be reasonable for accounts exhibiting high spending power and a preference for luxury goods. Hence, by incorporating the financial profiles of the sender and receiver as features into our prediction models, we can further refine the predictions to account for deviations in clients' financial habits based on the context of the transaction.

\begin{table}[!t]
\begin{center}
\caption{Risk of transaction format}
\begin{tabular}{|c | c| c |} 
\hline
\textbf{Format} & \textbf{Count} & \textbf{Probability of laundering given format} \\
\hline
Reinvestment & 0 & $0.00$ \\
Cheque & 324 & $1.7 \times 10^{-4}$ \\
Credit Card & 206 & $1.6 \times 10^{-4}$ \\
ACH & 4,483 & $7.5 \times 10^{-3}$ \\
Cash & 108 & $2.2 \times 10^{-4}$ \\
Wire & 0 & $0.00$ \\
Bitcoin & 56 & $3.8 \times 10^{-4}$ \\
\hline
\end{tabular}
\end{center}
\hfill \break
{\raggedright \small Note: These statistical insights in Tables II to \ref{table:volume} are generated through data analytics on the IT-AML dataset, which is modeled after real-life environments. Similar values can also be derived from global or geographical statistics obtained from domain studies or related research on the respective money laundering features. \par}
\label{table:format}
\end{table}

\begin{table}[!t]
\centering
\caption{Risk of transaction currency}
\begin{tabular}{ |l | c |} 
\hline
\textbf{Currency} & \textbf{Prob. of laundering given currency} \\ [1ex]
\hline
Australian Dollar & $7.9 \times 10^{-3}$ \\
Bitcoin & $4.0 \times 10^{-4}$ \\
Brazil Real & $6.4 \times 10^{-3}$ \\
Canadian Dollar & $7.3 \times 10^{-3}$ \\
Euro & $9.1 \times 10^{-3}$ \\
Mexican Peso & $7.1 \times 10^{-3}$ \\
Ruble & $7.3 \times 10^{-3}$ \\
Rupee & $7.1 \times 10^{-3}$ \\
Saudi Riyal & $3.8 \times 10^{-2}$ \\
Shekel & $4.3 \times 10^{-3}$ \\
Swiss Franc & $6.8 \times 10^{-3}$ \\
UK Pound & $5.4 \times 10^{-3}$ \\
US Dollar & $7.5 \times 10^{-3}$ \\
Yen & $7.9 \times 10^{-3}$ \\
Yuan & $5.4 \times 10^{-3}$ \\
\hline
\end{tabular}
\label{table:currency}
\vspace{-5mm}
\end{table}

\subsection{AML Risk Indicators}
Red flags serve as critical warning signs derived from domain expertise or learned from experience, alerting institutions to suspicious activities that warrant investigative attention. In recent years, many nations and financial institutions have published various sets of AML red flags to inform their stakeholders and help them avoid unintended participation in money laundering activities. For example, the FATF~\cite{fatf_virtual_assets} has compiled a comprehensive list of red flags derived from hundreds of case studies related to money laundering contributed by various jurisdictions over the years. This includes indicators such as unexplained or evasive explanations regarding the source of funds, the rationale behind the choice of payment method, unverified sources of high-risk transfers, and more. Thus, leveraging AML red flags as domain expertise could help us to optimize our machine learning models for a more accurate assessment of the risk that a given transaction is linked to money laundering.

\begin{table*}[!t]
\begin{center}
\caption{Risk of transaction frequency}
\begin{tabular}{||c | c c ||} 
\hline
\textbf{Transaction Frequency} & \textbf{Percentage of normal transaction} & \textbf{Percentage of money laundering transaction} \\ [1ex]
\hline
One-time & 0.20 & 0.80 \\
0-8 hours & 0.47 & 0.07 \\
8-24 hours & 0.14 & 0.11 \\
More than 24 hours & 0.19 & 0.02 \\ 
\hline
\end{tabular}
\end{center}
\hfill \break
{\raggedright \small Note: The transaction frequency refers to the time interval between subsequent transactions occurring between the same pair of sending and receiving accounts within the 10-day period recorded in the IT-AML dataset.}
\label{table:freq}
\end{table*}

    \begin{table*}[h!]
    \centering
    \caption{Risk of transaction volume given the amount in USD.}
     \begin{tabular}{||c | c c c c c c||} 
     \hline
    \textbf{Amount range in USD} & \textbf{100} & \textbf{1,000} & \textbf{10,000} & \textbf{25,000} & \textbf{100,000} & \textbf{above 100,000}\\ 
     \hline
     Percentage of normal transaction & $2.0 \times 10^{-1}$ & $3.2 \times 10^{-1}$ & $2.9 \times 10^{-1}$ & $7.0 \times 10^{-2}$ & $5.6 \times 10^{-2}$ & $6.4 \times 10^{-2}$ \\
     Percentage of money laundering transaction & $3.0 \times 10^{-2}$ & $1.2 \times 10^{-1}$ & $5.0 \times 10^{-1}$ & $3.0 \times 10^{-1}$ & $2.1 \times 10^{-2}$ & $2.9 \times 10^{-2}$ \\
     Prob. of laundering given trans. volume & $1.1 \times 10^{-4}$ & $4.1 \times 10^{-4}$ & $1.8 \times 10^{-3}$ & $4.4 \times 10^{-3}$ & $3.7 \times 10^{-4}$ & $4.5 \times 10^{-4}$ \\
     \hline
     \end{tabular}
     \label{table:volume}
     \vspace{-5mm}
    \end{table*}

In our model, we employed the following AML risk indicators derived from common AML red flags proposed by FATF's guidelines on virtual assets~\cite{fatf_virtual_assets}:
\begin{itemize}
    \item \textbf{Risk of transaction format:} Transactions that involve more than one type of payment format, especially those using anonymity-enhanced cryptocurrencies such as Bitcoin, are at a higher risk of money laundering compared to traditional wire transactions. This is due to differences in regulatory scrutiny and verification efforts. Cryptocurrencies and Automated Clearing House (ACH) transfers are more popular for money laundering because they can be conducted directly and immediately, whereas wire transactions often take up to two days to process and are closely monitored by the Office of Foreign Assets Control for international transfers for money laundering. 
    \item \textbf{Risk of transaction currency:} Currencies from nations with weak AML legislation often carry a higher risk of money laundering due to factors such as lower regulatory oversight, greater anonymity features, or instability in the financial system. For instance, the FATF evaluates nations based on their efforts to combat money laundering and terrorist financing according to a set of recommendations. Nations such as Mexico and Saudi Arabia have shown weaker compliance with these recommendations compared to countries such as Norway or China~\cite{FATF_countries}. 
    \item \textbf{Risk of transaction volume:} Transactions involving money laundering activities are generally larger in volume compared to normal transactions. For instance, 52\% of normal transactions and only 15\% of money laundering transactions in the dataset involve amounts less than 1,000 USD, while 36\% of normal transactions and 80\% of money laundering transactions fall within the range of 1,000 to 25,000 USD. Therefore, transactions exceeding a predefined threshold value, such as 1,000 USD, carry a higher risk of money laundering than transactions less than the threshold.
    \item \textbf{Risk of cross-bank and/or cross-currency transactions:} Transactions that occur across banks, denominated in the same payment and receiving currencies, are popular forms of money laundering transactions due to their lower traceability. This is attributed to the high volume of same-currency transactions and the restrictive data-sharing policies between banks, which render such transactions more challenging to detect. In our dataset, the probability of a transaction being money laundering is $2.0 \times 10^{-5}$ for same-bank transactions and $9.99 \times 10^{-4}$ for cross-bank transactions with the same currency. There are no cross-currency transactions in the dataset that involve money laundering, which aligns with the strict regulatory efforts that nations have implemented to monitor currency exchanges in the financial market.
    \item \textbf{Risk of transaction frequency:} An important indicator of scams and money laundering activities is a one-time, fast interaction with an unfamiliar account that the victim has never previously engaged with. For example, criminals can attempt to obscure the origin of funds by routing them through multiple intermediary accounts, making the funds difficult to trace.
\end{itemize}

In real-world scenarios, the values of AML risk indicators can be derived from global trends or adapted to specific banking regions for more robust predictions and even include other potentially important red flags. For this study, we derive domain-specific AML risk indicators by applying data exploration techniques to the transaction records in the dataset, as it closely mirrors real-world statistics. These risk indicator values are presented in Table II to \ref{table:volume}.

In our experiments, we compute the risk of money laundering for each transaction by evaluating the probability that it is associated with money laundering to enhance better AML predictions, taking into account transaction-specific risk features such as transaction format, currency, frequency, and amount in USD. For example, if the transaction is conducted in Saudi Riyal, the transaction has a 3.8\% risk in its transaction currency. These risk features are also incorporated into the prediction model alongside the actual transaction record. In addition, after the model generates a prediction, we employ these risk indicators as a final step to verify the prediction outcome by filtering the false positives. This approach helps reduce false alarms and enhances the explainability of our model's predictions.

\subsection{CRP-AML model}

\begin{table*}[ht!]
\begin{center}
\caption{Parameter tuning results on the sample dataset.}
\label{table:param}
\begin{tabular}{||c c c  c ||} 
\hline
\textbf{Number of encoder layers} & \textbf{Embedding size} & \textbf{Loss function} & \textbf{Minority class F1 scores}\\
\hline
1 & 64 & Binary cross-entropy loss & 0.7236  \\
2 & 32 & Binary focal cross-entropy loss & 0.731  \\
2 & 64 & Binary cross-entropy loss & 0.678  \\
2 & 64 & Binary focal cross-entropy loss & 0.759  \\
2 & 128 & Binary focal cross-entropy loss & \textbf{0.8345} \\
2 & 256 & Binary focal cross-entropy loss & 0.743 \\
3 & 64 & Binary cross-entropy loss & 0.709 \\
\hline
\end{tabular}
\end{center}
\vspace{-5mm}
\end{table*}

As discussed in previous sections, numerous approaches have been developed to detect money laundering activities in transactions, yet many remain inflexible or limited in their applicability to real-world settings. The situation is further complicated by the sheer volume of legitimate transactions conducted by the suspected account. Criminals may obscure illicit funds by mixing them with legitimate transactions during different stages of the money laundering process. This makes it challenging to identify a single illicit transaction among thousands of legitimate ones carried out for daily activities, without carefully examining the context of each transaction~\cite{10.1007/s10115-017-1144-z}.

To address these challenges, our framework aims to integrate various elements to improve money laundering detection under real-world constraints of limited data availability and privacy concerns. It utilize the following:
\begin{itemize}
    \item \textbf{Context:} The context of each transaction or the account profiles of the transacting parties~\cite{cui2021remember}, such as the identities of the transacting parties, their spending behaviors, and the financial relationships between them.
    \item \textbf{Risk:} The AML risks of the transaction from domain knowledge and statistical insights, such as the probability that criminals would use such a transaction for money laundering or use a specific currency for illicit activity~\cite{9328094}.
    \item \textbf{Predict:} The machine learning prediction model utilizes the context and risk indicators and is trained to adapt effectively to the evolving trends in money laundering and the specific characteristics of the dataset.
\end{itemize}

\begin{table*}[ht]
\centering
\caption{Minority class F1 scores (\%) for the datasets in IT-AML.}
\label{tab: modelresult}
\begin{tabular}{lccccc} 
\toprule
\textbf{Model} & \textbf{IT-AML Small High-Illicit} & \textbf{IT-AML Small Low-Illicit} & \textbf{IT-AML Med High-Illicit} & \textbf{IT-AML Med Low-Illicit} \\
\midrule
LightGBM+GFs~\cite{altman2023realistic} & 62.86 $\pm$ 0.25 & 20.83 $\pm$ 1.50 & 59.48 $\pm$ 0.15 & 20.85 $\pm$ 0.38 \\
XGBoost+GFs~\cite{altman2023realistic} & 63.23 $\pm$ 0.17 & 27.30 $\pm$ 0.33 & 65.70 $\pm$ 0.26 & 28.16 $\pm$ 0.14 \\
GIN~\cite{xu2018powerful,hu2020strategiespretraininggraphneural}& 28.70 $\pm$ 1.13 & 7.90 $\pm$ 2.78 & 42.20 $\pm$ 0.44 & 3.86 $\pm$ 3.62 \\
PNA~\cite{veličković2018deepgraphinfomax} & 56.77 $\pm$ 2.41 & 14.85 $\pm$ 1.46 & 59.71 $\pm$ 1.91 & 27.73 $\pm$ 1.65 \\
GIN+EU~\cite{battaglia2018relationalinductivebiasesdeep} & 47.73 $\pm$ 7.86 & 20.62 $\pm$ 2.41 & 49.26 $\pm$ 4.02 & 6.19 $\pm$ 8.32 \\
R-GCN~\cite{10.1007/978-3-319-93417-4_38} & 41.78 $\pm$ 0.48 & 7.42 $\pm$ 0.38 & Out-Of-Memory & Out-Of-Memory \\
GIN+EgoIDs~\cite{You_Gomes-Selman_Ying_Leskovec_2021} & 39.65 $\pm$ 4.73 & 14.98 $\pm$ 2.66 & 45.26 $\pm$ 2.16 & 11.17 $\pm$ 6.41 \\
GIN+Ports~\cite{sato2019approximationratiosgraphneural} & 54.85 $\pm$ 0.89 & 21.41 $\pm$ 2.40 & 54.22 $\pm$ 1.94 & 10.51 $\pm$ 12.82 \\
\midrule
GIN+ReverseMP~\cite{jaume2019edgnnsimplepowerfulgnn} & 46.79 $\pm$ 4.97 & 15.98 $\pm$ 4.39 & 51.93 $\pm$ 2.90 & 14.00 $\pm$ 9.34 \\
\ +Ports & 56.85 $\pm$ 2.44 & 23.80 $\pm$ 4.07 & 57.15 $\pm$ 0.76 & 11.39 $\pm$ 8.36 \\
\ +EgoIDs (Multi-GIN) & 57.15 $\pm$ 4.99 & 22.12 $\pm$ 2.88 & 56.23 $\pm$ 1.51 & 14.55 $\pm$ 2.91 \\
\midrule
Multi-GIN+EU~\cite{egressy2024provablypowerfulgraphneural} & 64.79 $\pm$ 1.22 & 26.88 $\pm$ 6.63 & 58.92 $\pm$ 1.83 & 16.30 $\pm$ 4.73 \\
Multi-PNA~\cite{egressy2024provablypowerfulgraphneural} & 64.59 $\pm$ 3.60 & 30.65 $\pm$ 2.00 & 65.67 $\pm$ 2.66 & 33.23 $\pm$ 1.31 \\
Multi-PNA+EU~\cite{egressy2024provablypowerfulgraphneural} & 68.16 $\pm$ 2.65 & 33.07 $\pm$ 2.63 & 66.48 $\pm$ 1.63 & 36.07 $\pm$ 1.17 \\
\midrule
\textbf{CRP-AML (Ours)}& 82.51 $\pm$ 2.62 & 60.55 $\pm$ 4.64 & 80.21 $\pm$ 1.92 & 62.17 $\pm$ 2.54 \\
\bottomrule
\end{tabular}

\hfill \break
{\raggedright \small High-illicit datasets have an illicit ratio of approximately 0.001. Medium-illicit datasets have an illicit transaction ratio ranging between 0.001 and 0.0005, while low-illicit datasets exhibit an even smaller ratio, with illicit transactions accounting for approximately 0.0005.\par}
\vspace{-5mm}
\end{table*}

With reference to Fig.\ref{fig:model}, the input to our prediction network consists of a combination of context embeddings from the sender and receiver account profiles, the original transaction record to be classified, and the risk features generated from the transaction and risk indicator values. These inputs are fed into the prediction network's input layer, which passes through a series of encoding modules, regularization layers, and decoding modules, generating a prediction at the output layer using a Sigmoid activation function. Once the prediction is made, the risk indicators are applied to filter out false positives, providing the final prediction for the transaction records.

\paragraph{\textbf{Network input}}
The input to the network consists of four components: (1) the original transaction record from the dataset; (2) risk features computed using the transaction record and AML risk indicators; (3) derived features from the transaction record, such as the time since the last transaction between the sender and receiver and the transaction categories; and (4) the transaction context, which includes context encodings derived from the transaction account profiles. In specific, the context encoding represents the following features and is trained using a simple RNN with two dense layers.
\begin{itemize}
    \item \textbf{Sender's or receiver's historical features}: Includes the total number of incoming and outgoing transactions, the average amount per incoming and outgoing transaction, the average time between subsequent transactions with the same transaction partner, the most frequent currency and format used, and the top 5 transaction types (e.g., medium-sized transactions conducted in USD via cheque and large-sized transactions in EUR via credit card).
    \item \textbf{Sender's or receiver's class features}: Captures the average number of extra large, large, medium and small transactions performed for this category of accounts; the average number of transactions in formats such as cheque, ACH, credit card, wire and bitcoin; and the average number of transactions conducted in currencies like USD, Swiss Franc, etc.
\end{itemize}
In addition, categorical features are one-hot encoded, and each input feature is normalized to stabilize training. The Tanh activation function is applied to the context encodings.

\paragraph{\textbf{Network parameters}} Random sampling is used to identify an appropriate set of hyperparameters, including the number of RNN layers in each encoding and decoding module, the size of hidden embeddings, the activation functions at different layers, the network's learning rate, and the parameters of the loss function. The range is then narrowed down to determine the optimal set of network parameters based on the best average validation score across training with five random starting seeds on the sample set. A summary of some of the parameter tuning results on the sample dataset is presented in Table~\ref{table:param}.

\paragraph{\textbf{Network architecture}}
The prediction network consists of an input layer, two encoding modules, a regularization layer, a decoding module, and a final prediction output layer. Each encoding module contains two RNN layers with embedding sizes of 128, using Tanh and ReLU activation functions, respectively, followed by a batch normalization layer. L2 regularization is then applied to the concatenated outputs of the encoding modules. The decoding module consists of two RNN layers with Leaky ReLU activation functions and an embedding size of 64. The output layer is a Dense layer with a sigmoid activation function and a size of 1. Adam optimizer is used with a learning rate of 0.001.

\paragraph{\textbf{Loss function}}
Since our dataset is extremely imbalanced, we choose binary focal cross-entropy loss~\cite{8417976} as our loss function to focus our training on the minority class and prevent the majority negative samples from overwhelming the cross-entropy loss. The loss function is defined as:

\begin{equation}
L(y, \hat{p}) = 
\begin{cases} 
-\alpha (1 - \hat{p})^{\gamma} \log(\hat{p}), & y = 1 \\
-(1 - \alpha) \hat{p}^{\gamma} \log(1 - \hat{p}), & \text{otherwise}
\end{cases}
\end{equation}

Here, $y \in \{0, 1\}$ represents a binary class label indicating whether a transaction involves money laundering, while $\hat{p} \in [0, 1]$ denotes the estimated probability of money laundering. The parameter $\gamma$, set to 3.0, controls how much higher-confidence correct predictions influence the overall loss, where higher $\gamma$ values down-weight majority class samples more rapidly. Finally, $\alpha$, set to 0.25, is a parameter that governs the trade-off between precision and recall by adjusting the weighting of errors for the positive class.

\paragraph{\textbf{Network training}}
Instead of using the 60-20-20 temporal train-validation-test split as proposed by the dataset authors \cite{egressy2024provablypowerfulgraphneural}, our framework encodes past behavior and account characteristics as input features, reducing the need to emphasize temporal information separately. Therefore, we use an 80-20 train-test split to train our model. Since the dataset is highly imbalanced to reflect real-world conditions, accuracy is no longer a meaningful metric for evaluating the model's performance \cite{egressy2024provablypowerfulgraphneural}. As a result, the model's performance is evaluated using F1 score on the minority class in the validation set, i.e., the positive money laundering samples.

\begin{figure}
    \centering
    \subfigure[]
    {
        \includegraphics[height=1.8in]{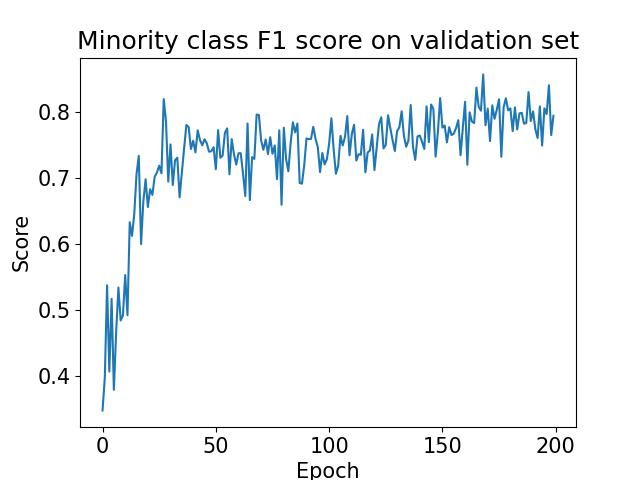}
        \label{fig:first_sub}
    }
    \\
    \subfigure[]
    {
        \includegraphics[height=1.8in]{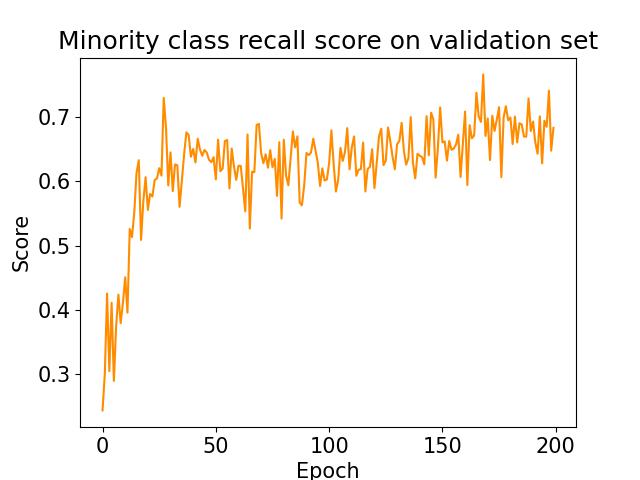}
        \label{fig:second_sub}
    }
    \caption{These figures show improvements of validation scores on the minority class as more epochs are run.}
    \label{fig:valscore}
    \vspace{-5mm}
\end{figure}

As an illustration of the training process, the improvements in minority F1 score and recall on the validation set with increasing training epochs are shown in Fig.~\ref{fig:valscore}. A sharp increase in performance is observed during the first 25 epochs, followed by gradual improvements, eventually stabilizing around 0.8 for the F1 score (i.e., Fig. 5a) and 0.7 for recall (Fig. 5b).

\paragraph{\textbf{Experimentation results}}
The experimental results and comparison to baseline models are presented in Table~\ref{tab: modelresult}. These results are obtained by averaging the validation outcomes across different datasets using five random seeds. Although the IT-AML dataset compilation includes additional datasets, our computational resources limited the processing of any larger datasets. 

As shown in Table~\ref{tab: modelresult}, the CRP-AML model outperforms pattern detection-based models, achieving the highest F1 score for the minority class at 82.51 $\pm$ 2.62, close to double that in other models when the ratio of money laundering records drops to 0.0005. In addition, applying the risk indicator filters to the prediction outcomes could improve precision by 2 to 5 percent depending on the initial seed and dataset used, effectively reducing false alarms. 

\subsection{Future Works}

\textcolor{black}{The experimental results demonstrate that our model significantly improves minority F1 scores, surpassing baseline models by over 10\% and achieving a best score of 82.51\% on the IT-AML dataset with 5 million transactions and an illicit ratio of 0.001 (IT-AML Small High-Illicit). However, while the model shows considerable improvements on the mid-sized dataset containing 32 million transaction records, its performance is notably lower compared to the smaller dataset. Due to computational resource constraints, we were unable to test our model on the IT-AML large dataset, which contains 180 million transactions and more closely reflects the transaction volumes typically observed in major international banks. Furthermore, the IT-AML dataset represents a cleaned dataset that excludes real-world challenges such as duplicate entries, missing values, and conflicting records, all of which could lead to a degradation in model performance. Thus, moving forward, we aim to collaborate with industry partners to evaluate our model on real-world banking datasets, thereby assessing its generalizability and robustness in practical scenarios.}

\textcolor{black}{In addition, our model aims to enhance explainability by highlighting the historical behaviors of both the sender and receiver, as well as the average behavior of individuals within similar categories. When combined with risk indicators and transaction records, these attributes allow for the identification and explanation of transactions that deviate significantly from the norm. However, many layering techniques are designed to inconspicuously blend illicit and benign transactions, making it challenging to detect and explain deviations when transaction attributes appear similar. In future work, we plan to incorporate additional domain knowledge from financial crime experts to develop systematic mechanisms for explaining the subtle distinctions between such transactions.}

\section{Conclusions} \label{sec: conclusion}
In summary, this paper offers a comprehensive review of deep learning solutions for AML, highlighting the strengths of various techniques in addressing AML challenges. Through a study of deep learning approaches such as CNN-based, RNN-based, Transformer-based, GNN-based, and DRL-based models, we identified several common challenges that hinder their feasibility and performance. For example, critical issues such as data quality, security, privacy, and regulatory compliance remain significant obstacles to the effective implementation of many of the reviewed proposals. 

\textcolor{black}{To address these challenges, we propose the CRP-AML model, which seeks to enhance the overall effectiveness of deep learning-based AML systems under real-world constraints, such as limited data availability, while safeguarding the privacy of innocent users.} Our approach incorporates the least-privilege principle by leveraging machine learning techniques, codifying AML red flags, and conducting account profiling to provide context-aware predictions for robust fraud detection despite data limitations. This approach have offered several benefits over existing AML techniques. For example, our CRP-AML model relies solely on existing internal banking records and information obtained through KYC procedures, eliminating the need to share external information with other banking institutions and ensuring compliance with banking privacy laws. In addition, our context-aware model provides individualized predictions, reducing the high false alarm rates often caused by over-generalization of AML metrics. Moreover, during AML investigations, only the account profile features and transactions related to the suspect and associated accounts are provided to AML teams, thereby minimizing the exposure of sensitive banking information. Finally, our model demonstrates significant improvements in the F1 score for the minority class compared to other models.

\bibliographystyle{bibstyle}
\bibliography{references}

\begin{thebibliography}{100}

\bibitem{MASCIANDARO1999}
D.~MASCIANDARO, Money Laundering: the Economics of Regulation {\em European
  Journal of Law and Economics}, vol.~7, pp.~225--240, May 1999.

\bibitem{usao2024moneylaundering}
M.~D. o.~F. United States Department~of Justice, Conspirators in Multi-Million
  Dollar International Money Laundering Conspiracy Sentenced 2024.
\newblock Accessed: 2024-12-04.

\bibitem{TechnologyandAntiMoneyLaundering}
D.~S. Demetis, {\em Technology and Anti-Money Laundering: A Systems Theory and
  Risk-Based Approach}.
\newblock Cheltenham, UK: Edward Elgar Publishing, 2010.

\bibitem{10506539}
R.~Zhang, K.~Xiong, H.~Du, D.~Niyato, J.~Kang, X.~Shen, and H.~V. Poor,
  Generative AI-Enabled Vehicular Networks: Fundamentals, Framework, and Case
  Study {\em IEEE Network}, vol.~38, no.~4, pp.~259--267, 2024.

\bibitem{thommandru2023smurfing}
A.~Thommandru, Smurfing in Electronic Banking: A Legal Investigation of the
  Potential for Transnational Money Laundering {\em International Journal of
  Legal Information}, vol.~51, no.~1, pp.~69--76, 2023.

\bibitem{albrecht2019use}
C.~Albrecht, K.~M. Duffin, S.~Hawkins, and V.~M. Morales~Rocha, The use of
  cryptocurrencies in the money laundering process {\em Journal of Money
  Laundering Control}, vol.~22, no.~2, pp.~210--216, 2019.

\bibitem{usdoj2024binance}
O.~o. P.~A. United States Department~of Justice, Binance and CEO Plead Guilty
  to Federal Charges in \$4B Resolution 2024.
\newblock Accessed: 2024-12-04.

\bibitem{fatf_virtual_assets}
FATF, ``Virtual assets: red flag indicators.'' FATF, 2022.

\bibitem{ISHIBUCHI20074}
H.~Ishibuchi and Y.~Nojima, Analysis of interpretability-accuracy tradeoff of
  fuzzy systems by multiobjective fuzzy genetics-based machine learning {\em
  International Journal of Approximate Reasoning}, vol.~44, no.~1, pp.~4--31,
  2007.
\newblock Genetic Fuzzy Systems and the Interpretability–Accuracy Trade-off.

\bibitem{westintell}
W.~Jarrod and B.~Maumita, Intelligent financial fraud detection: a
  comprehensive review {\em Computers \& security}, vol.~57, pp.~47--66, 2016.

\bibitem{shaikh2018model}
A.~K. Shaikh and A.~Nazir, A model for identifying relationships of suspicious
  customers in money laundering using social network functions in {\em
  Proceedings of the world congress on engineering}, vol.~1, pp.~4--7, 2018.

\bibitem{7838276}
E.~L. Paula, M.~Ladeira, R.~N. Carvalho, and T.~Marzagão, Deep Learning
  Anomaly Detection as Support Fraud Investigation in Brazilian Exports and
  Anti-Money Laundering in {\em 2016 15th IEEE International Conference on
  Machine Learning and Applications (ICMLA)}, pp.~954--960, 2016.

\bibitem{FARRUGIA2020113318}
S.~Farrugia, J.~Ellul, and G.~Azzopardi, Detection of illicit accounts over the
  Ethereum blockchain {\em Expert Systems with Applications}, vol.~150,
  p.~113318, 2020.

\bibitem{10058199}
J.~Fan, W.~Yang, Z.~Liu, J.~Kang, D.~Niyato, K.-Y. Lam, and H.~Du,
  Understanding Security in Smart City Domains From the ANT-Centric Perspective
  {\em IEEE Internet of Things Journal}, vol.~10, no.~13, pp.~11199--11223,
  2023.

\bibitem{10.1007/s10115-017-1144-z}
Z.~Chen, L.~D. Khoa, E.~N. Teoh, A.~Nazir, E.~K. Karuppiah, and K.~S. Lam,
  Machine learning techniques for anti-money laundering (AML) solutions in
  suspicious transaction detection: a review {\em Knowl. Inf. Syst.}, vol.~57,
  p.~245–285, Nov. 2018.

\bibitem{9446887}
D.~V. Kute, B.~Pradhan, N.~Shukla, and A.~Alamri, Deep Learning and Explainable
  Artificial Intelligence Techniques Applied for Detecting Money Laundering–A
  Critical Review {\em IEEE Access}, vol.~9, pp.~82300--82317, 2021.

\bibitem{moneylaunderingdetectionusingml}
J.~Alotibi, B.~Almutanni, T.~Alsubait, H.~Alhakami, and A.~Baz, Money
  Laundering Detection using Machine Learning and Deep Learning {\em
  International Journal of Advanced Computer Science and Applications},
  vol.~13, 01 2022.

\bibitem{10.1007/978-3-031-27762-7_8}
B.~Youssef, F.~Bouchra, and O.~Brahim, State of the Art Literature on
  Anti-money Laundering Using Machine Learning and Deep Learning Techniques in
  {\em The 3rd International Conference on Artificial Intelligence and Computer
  Vision (AICV2023), March 5--7, 2023} (A.~E. Hassanien, A.~Haqiq, A.~T. Azar,
  K.~Santosh, M.~A. Jabbar, A.~S{\l}owik, and P.~Subashini, eds.), (Cham),
  pp.~77--90, Springer Nature Switzerland, 2023.

\bibitem{10025710}
R.~I.~T. Jensen and A.~Iosifidis, Fighting Money Laundering With Statistics and
  Machine Learning {\em IEEE Access}, vol.~11, pp.~8889--8903, 2023.

\bibitem{GDPR_2022}
{The European Parliament and Council of the European}, ``General data
  protection regulation {(GDPR)}.'' \url{https://gdpr-info.eu/}, Sep 2022.

\bibitem{PIPL_2022}
{The Standing Committee of the National People’s Congress}, ``{Personal
  information protection law of the People’s Republic of China (PIPL)}.''
  \url{https://personalinformationprotectionlaw.com/}, May 2022.

\bibitem{bs2022}
U.~T. F. C.~E. Network, ``The bank secrecy act.''
  \url{https://www.fincen.gov/resources/statutes-and-regulations/bank-secrecy-act},
  May 2022.

\bibitem{Bankingact_2021}
{Monetary Authority of Singapore}, ``{Banking Act 1970}.''
  \url{https://sso.agc.gov.sg/Act/BA1970}, December 2021.

\bibitem{10.1145/373256.373257}
A.~Schaad, J.~Moffett, and J.~Jacob, The Role-Based Access Control System of a
  European Bank: A Case Study and Discussion in {\em Proceedings of the Sixth
  ACM Symposium on Access Control Models and Technologies}, SACMAT '01, (New
  York, NY, USA), p.~3–9, Association for Computing Machinery, 2001.

\bibitem{HSBC}
{HSBC}, ``Group structure.''
  \url{https://www.hsbc.com/investors/investing-in-hsbc/group-structure}, 2023.

\bibitem{altman2023realistic}
E.~Altman, B.~Egressy, J.~Blanuša, and K.~Atasu, Realistic Synthetic Financial
  Transactions for Anti-Money Laundering Models 2023.

\bibitem{10371347}
J.~Wu, D.~Lin, Q.~Fu, S.~Yang, T.~Chen, Z.~Zheng, and B.~Song, Toward
  Understanding Asset Flows in Crypto Money Laundering Through the Lenses of
  Ethereum Heists {\em IEEE Transactions on Information Forensics and
  Security}, vol.~19, pp.~1994--2009, 2024.

\bibitem{muller2007anti}
W.~H. Muller, C.~H. Kalin, and J.~G. Goldsworth, {\em Anti-money laundering:
  international law and practice}.
\newblock John Wiley \& Sons, 2007.

\bibitem{bellomarini2020rule}
L.~Bellomarini, E.~Laurenza, and E.~Sallinger, Rule-based Anti-Money Laundering
  in Financial Intelligence Units: Experience and Vision. {\em RuleML+ RR
  (Supplement)}, vol.~2644, pp.~133--144, 2020.

\bibitem{10756563}
J.~Fan, Z.~Liu, H.~Du, J.~Kang, D.~Niyato, and K.-Y. Lam, Improving Security in
  IoT-Based Human Activity Recognition: A Correlation-Based Anomaly Detection
  Approach {\em IEEE Internet of Things Journal}, pp.~1--1, 2024.

\bibitem{amldisclosure}
H.~Nobanee and N.~Ellili, Anti-money laundering disclosures and banks'
  performance {\em Journal of financial crime}, vol.~25, no.~1, pp.~95--108,
  2018.

\bibitem{Huang2015}
J.~Y. Huang, Effectiveness of US anti-money laundering regulations and HSBC
  case study {\em Journal of Money Laundering Control}, vol.~18, pp.~525--532,
  Jan 2015.

\bibitem{10.1016/j.eswa.2021.116429}
W.~Hilal, S.~A. Gadsden, and J.~Yawney, Financial Fraud: A Review of Anomaly
  Detection Techniques and Recent Advances {\em Expert Syst. Appl.}, vol.~193,
  May 2022.

\bibitem{10595068}
I.~D. Mienye and N.~Jere, Deep Learning for Credit Card Fraud Detection: A
  Review of Algorithms, Challenges, and Solutions {\em IEEE Access}, vol.~12,
  pp.~96893--96910, 2024.

\bibitem{goecks}
L.~Goecks, A.~Korzenowski, P.~Neto, D.~Souza, and T.~Mareth, Anti‐money
  laundering and financial fraud detection: A systematic literature review {\em
  Intelligent Systems in Accounting, Finance and Management}, vol.~29, 05 2022.

\bibitem{Soltani}
M.~Soltani, A.~Kythreotis, and A.~Roshanpoor, Two decades of financial
  statement fraud detection literature review; combination of bibliometric
  analysis and topic modeling approach {\em Journal of Financial Crime},
  vol.~30, 04 2023.

\bibitem{OZBAYOGLU2020106384}
A.~M. Ozbayoglu, M.~U. Gudelek, and O.~B. Sezer, Deep learning for financial
  applications : A survey {\em Applied Soft Computing}, vol.~93, p.~106384,
  2020.

\bibitem{10477569}
X.~Luo, X.~Han, W.~Zuo, X.~Wu, and W.~Liu, MLaD²: A Semi-Supervised Money
  Laundering Detection Framework Based on Decoupling Training {\em IEEE
  Transactions on Information Forensics and Security}, vol.~19, pp.~4518--4533,
  2024.

\bibitem{elliptic_dataset}
Elliptic, Elliptic Dataset: Cryptocurrency and Financial Crime 2019.
\newblock Accessed: 2024-12-24.

\bibitem{xia2022novel}
P.~Xia, Z.~Ni, H.~Xiao, X.~Zhu, and P.~Peng, A novel spatiotemporal prediction
  approach based on graph convolution neural networks and long short-term
  memory for money laundering fraud {\em Arabian Journal for Science and
  Engineering}, vol.~47, no.~2, pp.~1921--1937, 2022.

\bibitem{10570372}
B.~Qin, H.~Pan, Y.~Dai, X.~Si, X.~Huang, C.~Yuen, and Y.~Zhang, Machine and
  Deep Learning for Digital Twin Networks: A Survey {\em IEEE Internet of
  Things Journal}, vol.~11, no.~21, pp.~34694--34716, 2024.

\bibitem{Han2018NextGenAD}
J.~Han, U.~Barman, J.~Hayes, J.~Du, E.~Burgin, and D.~Wan, NextGen AML:
  Distributed Deep Learning based Language Technologies to Augment Anti Money
  Laundering Investigation in {\em Annual Meeting of the Association for
  Computational Linguistics}, 2018.

\bibitem{9490350}
S.~Singh, R.~Sulthana, T.~Shewale, V.~Chamola, A.~Benslimane, and B.~Sikdar,
  Machine-Learning-Assisted Security and Privacy Provisioning for Edge
  Computing: A Survey {\em IEEE Internet of Things Journal}, vol.~9, no.~1,
  pp.~236--260, 2022.

\bibitem{9739684}
J.~Bian, A.~A. Arafat, H.~Xiong, J.~Li, L.~Li, H.~Chen, J.~Wang, D.~Dou, and
  Z.~Guo, Machine Learning in Real-Time Internet of Things (IoT) Systems: A
  Survey {\em IEEE Internet of Things Journal}, vol.~9, no.~11, pp.~8364--8386,
  2022.

\bibitem{tatulli2023hamlet}
M.~P. Tatulli, T.~Paladini, M.~D’Onghia, M.~Carminati, and S.~Zanero, HAMLET:
  a transformer based approach for money laundering detection in {\em
  International Symposium on Cyber Security, Cryptology, and Machine Learning},
  pp.~234--250, Springer, 2023.

\bibitem{10679152}
R.~Zhang, H.~Du, Y.~Liu, D.~Niyato, J.~Kang, Z.~Xiong, A.~Jamalipour, and
  D.~In~Kim, Generative AI Agents With Large Language Model for Satellite
  Networks via a Mixture of Experts Transmission {\em IEEE Journal on Selected
  Areas in Communications}, vol.~42, no.~12, pp.~3581--3596, 2024.

\bibitem{10670196}
R.~Zhang, H.~Du, D.~Niyato, J.~Kang, Z.~Xiong, A.~Jamalipour, P.~Zhang, and
  D.~I. Kim, Generative AI for Space-Air-Ground Integrated Networks {\em IEEE
  Wireless Communications}, vol.~31, no.~6, pp.~10--20, 2024.

\bibitem{9272624}
A.~Uprety and D.~B. Rawat, Reinforcement Learning for IoT Security: A
  Comprehensive Survey {\em IEEE Internet of Things Journal}, vol.~8, no.~11,
  pp.~8693--8706, 2021.

\bibitem{mltechnique}
Z.~Chen, V.~K. Le~Dinh, E.~N. Teoh, A.~Nazir, K.~K. Ettikan, and S.~L. Kim,
  Machine learning techniques for anti-money laundering (AML) solutions in
  suspicious transaction detection: a review {\em Knowledge and Information
  Systems}, vol.~57, pp.~245--285, 11 2018.

\bibitem{10814854}
I.~Xavier, E.~Veiga, A.~M. N.~C. Ribeiro, R.~de~Paula~Monteiro, M.~Oliveira,
  R.~Amaro, and D.~Pinheiro, Towards the Identification of Money Laundering in
  Bank Transactions using Recurrent Neural Networks in {\em 2024 IEEE Latin
  American Conference on Computational Intelligence (LA-CCI)}, pp.~1--5, 2024.

\bibitem{10724384}
K.~K. Girish and B.~Bhowmik, Money Laundering Detection in Banking Transactions
  using RNNs and Hybrid Ensemble in {\em 2024 15th International Conference on
  Computing Communication and Networking Technologies (ICCCNT)}, pp.~1--7,
  2024.

\bibitem{sym16030378}
F.~Wan and P.~Li, A Novel Money Laundering Prediction Model Based on a Dynamic
  Graph Convolutional Neural Network and Long Short-Term Memory {\em Symmetry},
  vol.~16, no.~3, 2024.

\bibitem{financialfraudtransaction}
Y.~Chen and M.~Du, Financial Fraud Transaction Prediction Approach Based on
  Global Enhanced GCN and Bidirectional LSTM {\em Computational Economics},
  pp.~1--20, 11 2024.

\bibitem{9446893}
Z.~Chen, W.~M. Soliman, A.~Nazir, and M.~Shorfuzzaman, Variational Autoencoders
  and Wasserstein Generative Adversarial Networks for Improving the Anti-Money
  Laundering Process {\em IEEE Access}, vol.~9, pp.~83762--83785, 2021.

\bibitem{kute2024explainable}
D.~V. Kute, B.~Pradhan, N.~Shukla, and A.~Alamri, Explainable deep learning
  model for predicting money laundering transactions {\em International Journal
  on Smart Sensing and Intelligent Systems}, vol.~17, no.~1, 2024.

\bibitem{lundberg2017unifiedapproachinterpretingmodel}
S.~Lundberg and S.-I. Lee, A Unified Approach to Interpreting Model Predictions
  2017.

\bibitem{yu2024deeplearningcrossbordertransaction}
Q.~Yu, Z.~Xu, and Z.~Ke, Deep Learning for Cross-Border Transaction Anomaly
  Detection in Anti-Money Laundering Systems 2024.

\bibitem{du2018topologyadaptivegraphconvolutional}
J.~Du, S.~Zhang, G.~Wu, J.~M.~F. Moura, and S.~Kar, Topology Adaptive Graph
  Convolutional Networks 2018.

\bibitem{pareja2020evolvegcn}
A.~Pareja, G.~Domeniconi, J.~Chen, T.~Ma, T.~Suzumura, H.~Kanezashi, T.~Kaler,
  T.~Schardl, and C.~Leiserson, Evolvegcn: Evolving graph convolutional
  networks for dynamic graphs in {\em Proceedings of the AAAI conference on
  artificial intelligence}, vol.~34, pp.~5363--5370, 2020.

\bibitem{10.1162/neco.1997.9.8.1735}
S.~Hochreiter and J.~Schmidhuber, Long Short-Term Memory {\em Neural Comput.},
  vol.~9, p.~1735–1780, Nov. 1997.

\bibitem{cho2014learningphraserepresentationsusing}
K.~Cho, B.~van Merrienboer, C.~Gulcehre, D.~Bahdanau, F.~Bougares, H.~Schwenk,
  and Y.~Bengio, Learning Phrase Representations using RNN Encoder-Decoder for
  Statistical Machine Translation 2014.

\bibitem{Chawla2002}
N.~V. Chawla, K.~W. Bowyer, L.~O. Hall, and W.~P. Kegelmeyer, SMOTE: Synthetic
  Minority Over-sampling Technique {\em Journal of Artificial Intelligence
  Research}, vol.~16, p.~321–357, June 2002.

\bibitem{Chen2016}
T.~Chen and C.~Guestrin, XGBoost: A Scalable Tree Boosting System in {\em
  Proceedings of the 22nd ACM SIGKDD International Conference on Knowledge
  Discovery and Data Mining}, KDD ’16, p.~785–794, ACM, Aug. 2016.

\bibitem{10.1145/3534678.3539300}
J.~You, T.~Du, and J.~Leskovec, ROLAND: Graph Learning Framework for Dynamic
  Graphs in {\em Proceedings of the 28th ACM SIGKDD Conference on Knowledge
  Discovery and Data Mining}, KDD '22, (New York, NY, USA), p.~2358–2366,
  Association for Computing Machinery, 2022.

\bibitem{10.1007/s11063-022-10904-8}
I.~Alarab and S.~Prakoonwit, Graph-Based LSTM for Anti-money Laundering:
  Experimenting Temporal Graph Convolutional Network with Bitcoin Data {\em
  Neural Process. Lett.}, vol.~55, p.~689–707, June 2022.

\bibitem{arjovsky2017wassersteingan}
M.~Arjovsky, S.~Chintala, and L.~Bottou, Wasserstein GAN 2017.

\bibitem{han2022survey}
K.~Han, Y.~Wang, H.~Chen, X.~Chen, J.~Guo, Z.~Liu, Y.~Tang, A.~Xiao, C.~Xu,
  Y.~Xu, {\em et~al.}, A survey on vision transformer {\em IEEE transactions on
  pattern analysis and machine intelligence}, vol.~45, no.~1, pp.~87--110,
  2022.

\bibitem{10531073}
R.~Zhang, H.~Du, Y.~Liu, D.~Niyato, J.~Kang, S.~Sun, X.~Shen, and H.~V. Poor,
  Interactive AI With Retrieval-Augmented Generation for Next Generation
  Networking {\em IEEE Network}, vol.~38, no.~6, pp.~414--424, 2024.

\bibitem{xu2023illegal}
C.~Xu, S.~Zhang, L.~Zhu, X.~Shen, and X.~Zhang, Illegal Accounts Detection on
  Ethereum Using Heterogeneous Graph Transformer Networks in {\em International
  Conference on Information and Communications Security}, pp.~665--680,
  Springer, 2023.

\bibitem{lin2024fraudgt}
J.~Lin, X.~Guo, Y.~Zhu, S.~Mitchell, E.~Altman, and J.~Shun, FraudGT: A Simple,
  Effective, and Efficient Graph Transformer for Financial Fraud Detection in
  {\em Proceedings of the 5th ACM International Conference on AI in Finance},
  pp.~292--300, 2024.

\bibitem{9758694}
D.~Labanca, L.~Primerano, M.~Markland-Montgomery, M.~Polino, M.~Carminati, and
  S.~Zanero, Amaretto: An Active Learning Framework for Money Laundering
  Detection {\em IEEE Access}, pp.~1--1, 2022.

\bibitem{egressy2024provablypowerfulgraphneural}
B.~Egressy, L.~von Niederhäusern, J.~Blanusa, E.~Altman, R.~Wattenhofer, and
  K.~Atasu, Provably Powerful Graph Neural Networks for Directed Multigraphs
  2024.

\bibitem{LaundroGraph}
M.~Cardoso, P.~Saleiro, and P.~Bizarro, LaundroGraph: Self-Supervised Graph
  Representation Learning for Anti-Money Laundering in {\em Proceedings of the
  Third ACM International Conference on AI in Finance}, ICAIF '22, (New York,
  NY, USA), p.~130–138, Association for Computing Machinery, 2022.

\bibitem{10.1145/3409073.3409080}
I.~Alarab, S.~Prakoonwit, and M.~I. Nacer, Competence of Graph Convolutional
  Networks for Anti-Money Laundering in Bitcoin Blockchain in {\em Proceedings
  of the 2020 5th International Conference on Machine Learning Technologies},
  ICMLT '20, (New York, NY, USA), p.~23–27, Association for Computing
  Machinery, 2020.

\bibitem{weberanti}
W.~Mark, D.~Giacomo, C.~Jie, W.~D.~K. I, B.~Claudio, R.~Tom, and L.~C. E,
  Anti-money laundering in bitcoin: Experimenting with graph convolutional
  networks for financial forensics {\em arXiv preprint arXiv:1908.02591}, 2019.

\bibitem{lo2023inspection}
W.~W. Lo, G.~K. Kulatilleke, M.~Sarhan, S.~Layeghy, and M.~Portmann,
  Inspection-L: self-supervised GNN node embeddings for money laundering
  detection in bitcoin {\em Applied Intelligence}, vol.~53, no.~16,
  pp.~19406--19417, 2023.

\bibitem{10.1007/978-3-031-33377-4_10}
W.~Hyun, J.~Lee, and B.~Suh, Anti-Money Laundering in Cryptocurrency
  via Multi-Relational Graph Neural Network in {\em Advances in Knowledge
  Discovery and Data Mining} (H.~Kashima, T.~Ide, and W.-C. Peng, eds.),
  (Cham), pp.~118--130, Springer Nature Switzerland, 2023.

\bibitem{9332279}
J.~Wu, J.~Liu, W.~Chen, H.~Huang, Z.~Zheng, and Y.~Zhang, Detecting Mixing
  Services via Mining Bitcoin Transaction Network With Hybrid Motifs {\em IEEE
  Transactions on Systems, Man, and Cybernetics: Systems}, vol.~52, no.~4,
  pp.~2237--2249, 2022.

\bibitem{10114503}
D.~Cheng, Y.~Ye, S.~Xiang, Z.~Ma, Y.~Zhang, and C.~Jiang, Anti-Money Laundering
  by Group-Aware Deep Graph Learning {\em IEEE Transactions on Knowledge and
  Data Engineering}, vol.~35, no.~12, pp.~12444--12457, 2023.

\bibitem{oliveira2021guiltywalker}
C.~Oliveira, J.~Torres, M.~I. Silva, D.~Apar{\'\i}cio, J.~T. Ascens{\~a}o, and
  P.~Bizarro, GuiltyWalker: Distance to illicit nodes in the Bitcoin network
  {\em arXiv preprint arXiv:2102.05373}, 2021.

\bibitem{lightgbm}
G.~Ke, Q.~Meng, T.~Finley, T.~Wang, W.~Chen, W.~Ma, Q.~Ye, and T.-Y. Liu,
  LightGBM: A Highly Efficient Gradient Boosting Decision Tree 12 2017.

\bibitem{veličković2018deepgraphinfomax}
P.~Veličković, W.~Fedus, W.~L. Hamilton, P.~Liò, Y.~Bengio, and R.~D. Hjelm,
  Deep Graph Infomax 2018.

\bibitem{gasteiger2022predictpropagategraphneural}
J.~Gasteiger, A.~Bojchevski, and S.~Günnemann, Predict then Propagate: Graph
  Neural Networks meet Personalized PageRank 2022.

\bibitem{xu2018powerful}
K.~Xu, W.~Hu, J.~Leskovec, and S.~Jegelka, How powerful are graph neural
  networks? {\em arXiv preprint arXiv:1810.00826}, 2018.

\bibitem{hu2020strategiespretraininggraphneural}
W.~Hu, B.~Liu, J.~Gomes, M.~Zitnik, P.~Liang, V.~Pande, and J.~Leskovec,
  Strategies for Pre-training Graph Neural Networks 2020.

\bibitem{10.1007/978-3-319-93417-4_38}
M.~Schlichtkrull, T.~N. Kipf, P.~Bloem, R.~van den Berg, I.~Titov, and
  M.~Welling, Modeling Relational Data with Graph Convolutional Networks in
  {\em The Semantic Web} (A.~Gangemi, R.~Navigli, M.-E. Vidal, P.~Hitzler,
  R.~Troncy, L.~Hollink, A.~Tordai, and M.~Alam, eds.), (Cham), pp.~593--607,
  Springer International Publishing, 2018.

\bibitem{gat2018graph}
{Petar Veličković and Guillem Cucurull and Arantxa Casanova and Adriana
  Romero and Pietro Liò and Yoshua Bengio}, Graph Attention Networks 2018.

\bibitem{10.1145/3442381.3449989}
Y.~Liu, X.~Ao, Z.~Qin, J.~Chi, J.~Feng, H.~Yang, and Q.~He, Pick and Choose: A
  GNN-based Imbalanced Learning Approach for Fraud Detection in {\em
  Proceedings of the Web Conference 2021}, WWW '21, (New York, NY, USA),
  p.~3168–3177, Association for Computing Machinery, 2021.

\bibitem{hamilton2018inductiverepresentationlearninglarge}
W.~L. Hamilton, R.~Ying, and J.~Leskovec, Inductive Representation Learning on
  Large Graphs 2018.

\bibitem{NEURIPS2021_f1c15925}
C.~Ying, T.~Cai, S.~Luo, S.~Zheng, G.~Ke, D.~He, Y.~Shen, and T.-Y. Liu, Do
  Transformers Really Perform Badly for Graph Representation? in {\em Advances
  in Neural Information Processing Systems} (M.~Ranzato, A.~Beygelzimer,
  Y.~Dauphin, P.~Liang, and J.~W. Vaughan, eds.), vol.~34, pp.~28877--28888,
  Curran Associates, Inc., 2021.

\bibitem{10.1109/TSE.2024.3385538}
Y.~Li, X.~Dang, W.~Pian, A.~Habib, J.~Klein, and T.~F. Bissyand\'{e}, Test
  Input Prioritization for Graph Neural Networks {\em IEEE Trans. Softw. Eng.},
  vol.~50, p.~1396–1424, Apr. 2024.

\bibitem{10272628}
J.~Fan, M.~Xu, J.~Guo, L.~K. Shar, J.~Kang, D.~Niyato, and K.-Y. Lam,
  Decentralized Multimedia Data Sharing in IoV: A Learning-Based Equilibrium of
  Supply and Demand {\em IEEE Transactions on Vehicular Technology}, vol.~73,
  no.~3, pp.~4035--4050, 2024.

\bibitem{8102446}
A.~E. Bouchti, A.~Chakroun, H.~Abbar, and C.~Okar, Fraud detection in banking
  using deep reinforcement learning in {\em 2017 Seventh International
  Conference on Innovative Computing Technology (INTECH)}, pp.~58--63, 2017.

\bibitem{vimal2021applicationdeepreinforcementlearning}
S.~Vimal, K.~Kayathwal, H.~Wadhwa, and G.~Dhama, Application of Deep
  Reinforcement Learning to Payment Fraud 2021.

\bibitem{mnih2015human}
V.~Mnih, K.~Kavukcuoglu, D.~Silver, A.~A. Rusu, J.~Veness, M.~G. Bellemare,
  A.~Graves, M.~Riedmiller, A.~K. Fidjeland, G.~Ostrovski, {\em et~al.},
  Human-level control through deep reinforcement learning {\em nature},
  vol.~518, no.~7540, pp.~529--533, 2015.

\bibitem{DalPozzolo2015}
A.~D. Pozzolo, O.~Caelen, R.~A. Johnson, and G.~Bontempi, Calibrating
  probability with undersampling for unbalanced classification in {\em 2015
  IEEE Symposium Series on Computational Intelligence}, pp.~159--166, IEEE,
  2015.

\bibitem{kaggle_ieee_fraud_detection}
I.~C.~I. Society, IEEE-CIS Fraud Detection Competition 2019.
\newblock Accessed: 2025-01-05.

\bibitem{10.1007/s42979-021-00592-x}
I.~H. Sarker, Machine Learning: Algorithms, Real-World Applications and
  Research Directions {\em SN Comput. Sci.}, vol.~2, Mar. 2021.

\bibitem{jordan2015machine}
M.~I. Jordan and T.~M. Mitchell, Machine learning: Trends, perspectives, and
  prospects {\em Science}, vol.~349, no.~6245, pp.~255--260, 2015.

\bibitem{wang1996beyond}
R.~Y. Wang and D.~M. Strong, Beyond accuracy: What data quality means to data
  consumers {\em Journal of management information systems}, vol.~12, no.~4,
  pp.~5--33, 1996.

\bibitem{9964052}
Z.~Li, Y.~Zhang, Q.~Wang, and S.~Chen, Transactional Network Analysis and Money
  Laundering Behavior Identification of Central Bank Digital Currency of China
  {\em Journal of Social Computing}, vol.~3, no.~3, pp.~219--230, 2022.

\bibitem{OZTAS2024161}
B.~Oztas, D.~Cetinkaya, F.~Adedoyin, M.~Budka, G.~Aksu, and H.~Dogan,
  Transaction monitoring in anti-money laundering: A qualitative analysis and
  points of view from industry {\em Future Generation Computer Systems},
  vol.~159, pp.~161--171, 2024.

\bibitem{johnson2019survey}
J.~M. Johnson and T.~M. Khoshgoftaar, Survey on deep learning with class
  imbalance {\em Journal of big data}, vol.~6, no.~1, pp.~1--54, 2019.

\bibitem{10486886}
M.~R. Karim, F.~Hermsen, S.~A. Chala, P.~De~Perthuis, and A.~Mandal, Scalable
  Semi-Supervised Graph Learning Techniques for Anti Money Laundering {\em IEEE
  Access}, vol.~12, pp.~50012--50029, 2024.

\bibitem{vogelsgesang2018get}
A.~Vogelsgesang, M.~Haubenschild, J.~Finis, A.~Kemper, V.~Leis,
  T.~M{\"u}hlbauer, T.~Neumann, and M.~Then, Get real: How benchmarks fail to
  represent the real world in {\em Proceedings of the Workshop on Testing
  Database Systems}, pp.~1--6, 2018.

\bibitem{10496993}
S.~Deng, Z.~Lei, G.~Wen, Z.~S. Li, Y.~Zhang, K.~Feng, and X.~Chen, Knowledge
  Distillation-Guided Cost-Sensitive Ensemble Learning Framework for Imbalanced
  Fault Diagnosis {\em IEEE Internet of Things Journal}, vol.~11, no.~13,
  pp.~23110--23122, 2024.

\bibitem{smith2014instance}
M.~R. Smith, T.~Martinez, and C.~Giraud-Carrier, An instance level analysis of
  data complexity {\em Machine learning}, vol.~95, pp.~225--256, 2014.

\bibitem{9763022}
B.~A. Salau, A.~Rawal, and D.~B. Rawat, Recent Advances in Artificial
  Intelligence for Wireless Internet of Things and Cyber–Physical Systems: A
  Comprehensive Survey {\em IEEE Internet of Things Journal}, vol.~9, no.~15,
  pp.~12916--12930, 2022.

\bibitem{thommandru2023exploring}
A.~Thommandru, V.~Mone, S.~Mitharwal, and R.~Tilwani, Exploring the
  intersection of machine learning, money laundering, data privacy, and law in
  {\em 2023 International Conference on Innovative Data Communication
  Technologies and Application (ICIDCA)}, pp.~149--155, IEEE, 2023.

\bibitem{oagccpa}
{California Department of Justice}, California Consumer Privacy Act (CCPA)
  2024.
\newblock Accessed: 2024-12-31.

\bibitem{9830997}
Z.~Liu, J.~Guo, W.~Yang, J.~Fan, K.-Y. Lam, and J.~Zhao, Privacy-Preserving
  Aggregation in Federated Learning: A Survey {\em IEEE Transactions on Big
  Data}, pp.~1--20, 2022.

\bibitem{amlandterrorism}
N.~Al-Suwaidi and H.~Nobanee, Anti-money laundering and anti-terrorism
  financing: a survey of the existing literature and a future research agenda
  {\em Journal of Money Laundering Control}, vol.~ahead-of-print, 05 2020.

\bibitem{gill2004preventing}
M.~Gill and G.~Taylor, Preventing money laundering or obstructing business?
  Financial companies' perspectives on ‘know your customer’procedures {\em
  British Journal of Criminology}, vol.~44, no.~4, pp.~582--594, 2004.

\bibitem{10356200}
B.~Oztas, D.~Cetinkaya, F.~Adedoyin, M.~Budka, H.~Dogan, and G.~Aksu,
  Perspectives from Experts on Developing Transaction Monitoring Methods for
  Anti-Money Laundering in {\em 2023 IEEE International Conference on
  e-Business Engineering (ICEBE)}, pp.~39--46, 2023.

\bibitem{zhou2021evaluating}
J.~Zhou, A.~H. Gandomi, F.~Chen, and A.~Holzinger, Evaluating the quality of
  machine learning explanations: A survey on methods and metrics {\em
  Electronics}, vol.~10, no.~5, p.~593, 2021.

\bibitem{app12199637}
A.~Ali, S.~Abd~Razak, S.~H. Othman, T.~A.~E. Eisa, A.~Al-Dhaqm, M.~Nasser,
  T.~Elhassan, H.~Elshafie, and A.~Saif, Financial Fraud Detection Based on
  Machine Learning: A Systematic Literature Review {\em Applied Sciences},
  vol.~12, no.~19, 2022.

\bibitem{10021108}
R.~Searle, P.~Gururaj, A.~Gupta, and K.~Kannur, Secure Implementation of
  Artificial Intelligence Applications for Anti-Money Laundering using
  Confidential Computing in {\em 2022 IEEE International Conference on Big Data
  (Big Data)}, pp.~3092--3098, 2022.

\bibitem{10509682}
T.~Awosika, R.~M. Shukla, and B.~Pranggono, Transparency and Privacy: The Role
  of Explainable AI and Federated Learning in Financial Fraud Detection {\em
  IEEE Access}, vol.~12, pp.~64551--64560, 2024.

\bibitem{gilpin2018explaining}
L.~H. Gilpin, D.~Bau, B.~Z. Yuan, A.~Bajwa, M.~Specter, and L.~Kagal,
  Explaining explanations: An overview of interpretability of machine learning
  in {\em 2018 IEEE 5th International Conference on data science and advanced
  analytics (DSAA)}, pp.~80--89, IEEE, 2018.

\bibitem{10671571}
Q.~Wang, W.-T. Tsai, and T.~Shi, GraphALM: Active Learning for Detecting Money
  Laundering Transactions on Blockchain Networks {\em IEEE Network}, pp.~1--1,
  2024.

\bibitem{EDGE2009381}
M.~E. Edge and P.~R. {Falcone Sampaio}, A survey of signature based methods for
  financial fraud detection {\em Computers \& Security}, vol.~28, no.~6,
  pp.~381--394, 2009.

\bibitem{7277068}
X.~Ruan, Z.~Wu, H.~Wang, and S.~Jajodia, Profiling Online Social Behaviors for
  Compromised Account Detection {\em IEEE Transactions on Information Forensics
  and Security}, vol.~11, no.~1, pp.~176--187, 2016.

\bibitem{cui2021remember}
J.~Cui, C.~Yan, and C.~Wang, ReMEMBeR: Ranking metric embedding-based
  multicontextual behavior profiling for online banking fraud detection {\em
  IEEE Transactions on Computational Social Systems}, vol.~8, no.~3,
  pp.~643--654, 2021.

\bibitem{8735623}
N.~Kasa, A.~Dahbura, C.~Ravoori, and S.~Adams, Improving Credit Card Fraud
  Detection by Profiling and Clustering Accounts in {\em 2019 Systems and
  Information Engineering Design Symposium (SIEDS)}, pp.~1--6, 2019.

\bibitem{FATF_countries}
{FATF - Financial Action Task Force}, {Countries - FATF - Financial Action Task
  Force} 2024.
\newblock Accessed: June 8, 2024.

\bibitem{9328094}
M.~Alkhalili, M.~H. Qutqut, and F.~Almasalha, Investigation of Applying Machine
  Learning for Watch-List Filtering in Anti-Money Laundering {\em IEEE Access},
  vol.~9, pp.~18481--18496, 2021.

\bibitem{battaglia2018relationalinductivebiasesdeep}
P.~W. Battaglia, J.~B. Hamrick, V.~Bapst, A.~Sanchez-Gonzalez, V.~Zambaldi,
  M.~Malinowski, A.~Tacchetti, D.~Raposo, A.~Santoro, R.~Faulkner, C.~Gulcehre,
  F.~Song, A.~Ballard, J.~Gilmer, G.~Dahl, A.~Vaswani, K.~Allen, C.~Nash,
  V.~Langston, C.~Dyer, N.~Heess, D.~Wierstra, P.~Kohli, M.~Botvinick,
  O.~Vinyals, Y.~Li, and R.~Pascanu, Relational inductive biases, deep
  learning, and graph networks 2018.

\bibitem{You_Gomes-Selman_Ying_Leskovec_2021}
J.~You, J.~M. Gomes-Selman, R.~Ying, and J.~Leskovec, Identity-aware Graph
  Neural Networks {\em Proceedings of the AAAI Conference on Artificial
  Intelligence}, vol.~35, pp.~10737--10745, May 2021.

\bibitem{sato2019approximationratiosgraphneural}
R.~Sato, M.~Yamada, and H.~Kashima, Approximation Ratios of Graph Neural
  Networks for Combinatorial Problems 2019.

\bibitem{jaume2019edgnnsimplepowerfulgnn}
G.~Jaume, A.~phi Nguyen, M.~R. Martínez, J.-P. Thiran, and M.~Gabrani, edGNN:
  a Simple and Powerful GNN for Directed Labeled Graphs 2019.

\bibitem{8417976}
T.-Y. Lin, P.~Goyal, R.~Girshick, K.~He, and P.~Dollár, Focal Loss for Dense
  Object Detection {\em IEEE Transactions on Pattern Analysis and Machine
  Intelligence}, vol.~42, no.~2, pp.~318--327, 2020.

\end{thebibliography}

\end{document}